\documentclass[10pt,journal,compsoc]{IEEEtran}
\usepackage{amsmath,graphicx,amsfonts}
\usepackage{textcomp,float}
\usepackage{tablefootnote}
\usepackage{booktabs}
\usepackage{multirow}
\usepackage{enumitem}

\ifCLASSOPTIONcompsoc
  \usepackage[nocompress]{cite}
\else
  \usepackage{cite}
\fi
\ifCLASSINFOpdf
\else
\fi
\begin{document}
%
\title{Estimating the Uncertainty in Emotion Class Labels with Utterance-Specific Dirichlet Priors}
%
%
%
%

\author{Wen~Wu,\textit{ Student Member, IEEE}, Chao~Zhang,\textit{ Member, IEEE}, Xixin~Wu,\textit{ Member, IEEE}, and Philip~C.~Woodland, \textit{Fellow, IEEE}
\IEEEcompsocitemizethanks{
\IEEEcompsocthanksitem W. Wu and P. Woodland are with the Department of Engineering, University of Cambridge, Trumpington St., Cambridge. \\
Email: \{ww368, pcw\}@eng.cam.ac.uk
\IEEEcompsocthanksitem C. Zhang is with the Department of Electronic Engineering, Tsinghua University, Beijing, China. Work done while with Cambridge University. \\
Email: cz277@tsinghua.edu.cn
\IEEEcompsocthanksitem  X. Wu is with the Stanley Ho Big Data Decision Analytics Research Centre, The Chinese University of Hong Kong, Hong Kong SAR, China.\\
Email: wuxx@se.cuhk.edu.hk
}}


%
%

\markboth{Submitted to IEEE Transactions on Affective Computing,~2022}
{Wu \MakeLowercase{\textit{et al.}}: Estimating the Uncertainty in Emotion Class Labels with Utterance-Specific Dirichlet Priors}
%



\IEEEtitleabstractindextext{%
\begin{abstract}
Emotion recognition is a key attribute for artificial intelligence systems that need to naturally interact with humans. However, the task definition is still an open problem due to the inherent ambiguity of emotions. In this paper, a novel Bayesian training loss based on per-utterance Dirichlet prior distributions is proposed for verbal emotion recognition, which models the uncertainty in one-hot labels created when human annotators assign the same utterance to different emotion classes.
An additional metric is used to evaluate the performance by detecting test utterances with high labelling uncertainty. This removes a major limitation that emotion classification systems only consider utterances with labels where the majority of annotators agree on the emotion class. 
Furthermore, a frequentist approach is studied to leverage the continuous-valued ``soft'' labels obtained by averaging the one-hot labels.
We propose a two-branch model structure for emotion classification on a per-utterance basis, which achieves state-of-the-art classification results on the widely used IEMOCAP dataset.
Based on this, uncertainty estimation experiments were performed. 
The best performance in terms of the area under the precision-recall curve when detecting utterances with high uncertainty was achieved by interpolating the Bayesian training loss with the Kullback-Leibler divergence training loss for the soft labels. The generality of the proposed approach was verified using the MSP-Podcast dataset which yielded the same pattern of results.
\end{abstract}

}

\maketitle

\IEEEdisplaynontitleabstractindextext

%
\IEEEpeerreviewmaketitle

\IEEEraisesectionheading{\section{Introduction}\label{sec:introduction}}

%
%
%
%

\IEEEPARstart{A}{utomatic}
emotion recognition (AER), as a key component of affective computing, has attracted much attention due to its wide range of potential applications in conversational AI, driver monitoring, and mental health analysis \textit{etc.} Despite  significant progress in recent years \cite{Kim2013,Poria2017,Tzirakis2017}, AER is still a challenging problem and  does not even have a single widely accepted task definition.

A straightforward definition of AER is to classify each utterance (\textit{i.e.} a data sample of emotions with verbal evidence) into a set of pre-defined discrete emotional states (referred to as emotion classes in this paper, such as happy, sad, and frustrated \textit{etc.})~\cite{EMODB,CHEAVD,EmoVox,Busso2008IEMOCAPIE,EmoTV1,MSP-IMPROV,MSP-podcast,CMU-MOSEI}. However, since emotion is inherently ambiguous, complex, and highly personal, disagreements exist among human annotators who perceive the emotions and then label the data. Based on subjective judgements, different one-hot labels can be assigned to the same utterance by different annotators, and even by the same annotator~\cite{Busso2008IEMOCAPIE}, which leads to uncertainty in emotion labelling.  
Statistics show that some emotion classes are positively correlated (\textit{e.g.} sad \& frustrated) or inversely correlated (\textit{e.g.} sad \& happy)~\cite{Busso2008IEMOCAPIE,MSP-IMPROV,Meld}. These issues make the labels created by single annotators less reliable. As a solution, multiple annotators are often employed to label each utterance and majority voting is used to remove the uncertainty arising from difference among the assigned labels. 
However, this strategy not only causes training utterances without majority agreed labels to be unused, which is not ideal since emotional data are highly valuable, but also effectively assumes that such utterances would either not be encountered or not be evaluated at test-time. In fact, mixtures of emotions are commonly observed in human interaction~\cite{devillers2005challenges}.

In this paper, AER with  one-hot labels is studied and the classification-based task definition revisited. Instead of removing the uncertainty in labels with majority voting, we propose modelling such uncertainty with a novel Bayesian training loss based on utterance-specific Dirichlet prior distributions.
In this approach, the one-hot labels provided by different annotators of each utterance are considered as categorical distributions sampled from an utterance-specific Dirichlet prior distribution.
A separate prior distribution is estimated for each utterance, which is achieved by an improved Dirichlet prior network (DPN)~\cite{malinin2018predictive}. The DPN is trained by minimising the negative log likelihood of sampling the original one-hot labels from their relevant utterance-specific Dirichlet priors.
Alternatively, from a frequentist perspective,
``soft''  labels can be obtained by averaging the categorical distributions relevant to the one-hot labels which can be viewed as maximum likelihood estimate (MLE) of the label for each utterance. An AER model can be trained by minimising the Kullback-Leibler (KL) divergence between the soft labels and its output distributions. 
The DPN and KL divergence training losses can be combined by  simple linear interpolation.

To evaluate the proposed approach, a state-of-the-art neural network model architecture proposed in~\cite{wu2021emotion} is adopted for emotion classification, which consists of a time synchronous branch (TSB) that focuses on modelling the temporal correlations of multimodal features and a time asynchronous branch (TAB) that takes sentence embeddings as input to facilitate modelling meanings embedded in the text transcriptions. Experimental results on the widely used IEMOCAP dataset~\cite{Busso2008IEMOCAPIE} show that the TSB-TAB structure achieves state-of-the-art classification results in 4-way classification (happy, sad, angry \& neutral) when evaluated with all of the commonly used speaker-independent test setups. The 4-way classification is then extended to 5-way classification by including an extra emotion class ``others'' which represents all the other types of emotion labelled in IEMOCAP but ignored in the 4-way setup. Next, we redefine the task from classification to distribution modelling by representing the emotions using a 5-dimensional distribution rather than a single hard label. This allows utterances without majority agreed labels to also be considered. Uncertainty estimation is performed by training the model with soft labels and the DPN training loss. Classification accuracy is no longer an appropriate evaluation metric when considering uncertainty in emotion labelling. Instead, we propose evaluating the model performance in uncertainty estimation in terms of the area under the precision-recall curve (AUPR) when detecting utterances without majority unique labels at test-time. This also provides a more general two-step test procedure for AER, which can detect utterances with high uncertainty in emotion for further processing and classify the remainder of them into one of the emotion classes. 
Further experiments on the MSP-Podcast dataset, a larger speech corpus with natural emotions, demonstrate that our proposed uncertainty estimation approach generalises to handling  realistic emotion data.

The rest of the paper is organised as follows. Section~\ref{sec:relatedwork} discusses work on objective emotion quantification and related methods that address the inconsistency of emotion perception among human annotators. Section~\ref{sec:classification} presents an analysis of the IEMOCAP database and revisits the definition of emotion classification task with IEMOCAP. Section~\ref{sec:modelling} introduces the soft-label-based and DPN-based approaches to emotion distribution modelling and uncertainty analysis. The model structure and experiment setup are shown in Section~\ref{sec:expsetup}. Section~\ref{sec: classification exp} presents the results and analysis for emotion classification and distribution modelling on IEMOCAP. The proposed approach is further verified with experiments on the MSP-Podcast dataset in Section~\ref{sec: MSP}, followed by conclusions.

\section{Related work}
\label{sec:relatedwork}

The inherent ambiguity of emotion, resulting from mixed emotions and personal variations in emotion expression \textit{etc}., makes it still an open question as how to define emotion for easier quantification and objective analysis. 
Discrete emotion theory classifies emotion into several basic categories (\textit{{e.g.}} happy, sad, fear, anger \textit{etc.})~\cite{tomkins1962affect,ekman1992facial,ekman2004emotions} while psychologists have also observed that these distinct emotion categories overlap and have blurred boundaries between them~\cite{fehr1984concept,cowen2017self}. Alternative methods were developed to characterise emotional states by several fundamental continuous-valued or multi-valued bipolar dimensions that are more suitable to be evaluated independently ~\cite{russell1980circumplex,grimm2007primitives}. Despite the commonly assumed dimensions such as \textit{valence-arousal} and \textit{approach–avoidance}~\cite{russell1980circumplex,grimm2007primitives}, there's still a debate about the proper dimensional scheme and the orthogonality of the dimensions~\cite{schneirla1959evolutionary,lang1997motivated,watson1999two}. 

The subjectivity of emotional perception further complicates the problem of designing AER datasets. Despite the efforts of psychologists to de-correlate emotion dimensions, creating intensity labels with continuous values or multiple discrete values can still be highly subjective and also lead to a high degree of uncertainty in the data. 
In response to this problem, most datasets were created using the strategy of having multiple human annotators to
provide multiple labels for each utterance. The ``ground truth'' is then commonly defined as the majority vote for discrete labels~\cite{Busso2008IEMOCAPIE, CHEAVD2.0,MSP-IMPROV,Meld} or the mean for dimensional labels~\cite{RECOLA,Busso2008IEMOCAPIE,MSP-IMPROV}. 
When using the mean dimensional labels, the discrepancies between annotators are ignored. Several approaches have been proposed to characterise the subjective property of emotion perception by  modelling the inter-annotator disagreement level as the standard deviation of the dimensional labels, such as including a separate task to predict the standard deviation in a multi-task framework~\cite{han2021exploring,han2017hard}, or predicting such values using Gaussian mixture regression models~\cite{dang2017investigation,dang2018dynamic}. Recently, alternative methods including Gaussian processes~\cite{atcheson2019using}, generative variational auto-encoders~\cite{sridhar2021generative}, and Monte-Carlo dropout~\cite{sridhar2020modeling} have  been applied to the problem without  using the standard deviation of dimensional emotion labels as additional training labels.

When using a majority vote to obtain the ground truth for discrete class labelling, the data without ground truth due to annotators' disagreement are usually discarded in classification-based AER~\cite{tripathi2018multimodal,Majumder2018,Poria2018}. AER researchers have proposed various methods to address the uncertainty
in emotion labelling caused by the inconsistency of emotion perception among human annotators. Nediyanchath~\textit{et al.}~\cite{Nediyanchath2020} used multitask learning with gender or speaker classification to model the variations in personal aspects of emotional expression. Lotfian~\textit{et al.}~\cite{Lotfian_2018} proposed a multitask learning framework to recognise the primary emotional class by leveraging extra information provided in the evaluations about secondary emotions.
Ando~\textit{et al.}~\cite{Ando_2019} estimated the existence of multi-label emotions as an auxiliary task to improve the recognition of the dominant emotions. Another type of commonly used method is to train AER models with soft labels, which are derived as the mean of the hard labels and can be interpreted as the intensities of the emotion classes~\cite{Ando_2018}.
Fayek \textit{et al.}~\cite{Fayek_2016} incorporated inter-annotator variabilities by training a separate model based on the hard labels produced by each annotator, and it was shown that an ensemble of such models performed similarly to a single model trained using the soft labels. 
Although these approaches improved training with soft labels, at test-time the evaluations were only based on emotion classification accuracy, which results in a major inconsistency between training and evaluation~\cite{Mower09interpretingambiguous}. 

This paper focuses on classification-based AER. Rather than trying to remove the uncertainty in the emotion representation or to make emotion classes more separable, we acknowledge the ambiguity in emotion expression and subjectivity in emotion perception and evaluations, and model the resulted uncertainty in emotion class labels using a novel Bayesian training loss.

\section{Emotion Classification with IEMOCAP}
\label{sec:classification}

The IEMOCAP~\cite{Busso2008IEMOCAPIE} corpus is the primary corpus used in this paper. It is one of the most widely used datasets for verbal emotion classification and is designed with a typical data annotation procedure. It consists of 5 dyadic conversational sessions performed by 10 professional actors and the data includes three different modalities including the spoken audio, text transcription and the facial movements. There are in total 10,039 utterances and approximately 12 hours of data, with an average duration of 4.5 seconds per utterance.
Each utterance was annotated by three human annotators for categorical labels (neutral, happiness, sadness, and anger \textit{etc.}). Each annotator was allowed to tag more than one emotion category for each sentence if they perceived a mixture of emotions. The \textit{ground truth labels} are determined by majority voting. 
However, since only 7,532 utterances have majority unique\footnote{Emotion category with highest votes was unique (notice that the evaluators were
allowed to tag more than one emotion category). } hard labels , the remaining 25\% of the utterances are normally discarded.  
This issue and its solutions will be discussed later in detail in Sections~\ref{sec: re-exam} and \ref{sec:modelling}. 
Although our analysis is primarily performed on IEMOCAP, the issues also existed in many other commonly used AER  datasets,
such as MSP-Podcast \cite{MSP-podcast} whose analyses are provided in Section~\ref{sec: MSP}.

\subsection{4-way classification}
\label{ssec:4way}
To be consistent and comparable with previous studies \cite{Majumder2018,Kim2013,tripathi2018multimodal,Poria2018,liu20b_interspeech,chen20b_interspeech,makiuchi2021multimodal}, only utterances with majority unique labels
belonging to ``angry", ``happy", ``excited", ``sad", and ``neutral'' were used for our 4-way classification experiments. 
The ``excited'' class was merged with ``happy'' to better balance the size of each emotion class, which results in a total of 5,531 utterances 
with 1,636, 1,103, 1,084, and 1,708 utterances for happy, angry, sad and neutral respectively. 
This approach is widely used but it discards 44\% of the data.

\subsection{5-way classification}
\label{sec: 5-way classification}
As shown in Table~\ref{fig: groud-truth-distr}, although 7,532 utterances in IEMOCAP have majority unique labels, only 5,531 of them are used in the 4-way classification setup given in Section~\ref{ssec:4way}.  
This excludes all the other classes of emotions, including ``frustration'' which is the largest emotion class taking 25\% of the dataset, and therefore this partition only tackles part of the AER problem.  
To resolve this issue, we investigated the use of an alternative 5-way classification setup. An extra target of ``others'' was included as the 5-th class, to represent all of the other emotions that exist in IEMOCAP including utterances labelled as ``frustration", ``fear'', ``surprise'', `disgust'', and ``other''. 
All 7,532 utterances with majority unique labels were used for training and test in 5-way classification.

\begin{table}[htbp!] 
\centering    
\caption{Number of utterances associated with different classes of ground truth (majority unique) labels in IEMOCAP. The emotion class ``excited'' is merged into ``happy''. }

\scalebox{1}{
\begin{tabular}{ccccc}
\toprule
Anger    & Happiness & Sadness & Neutral & Frustration \\
1,103     & 1,636      & 1,084    & 1,708    & 1,849        \\
\midrule
\midrule
Surprise & Fear      & Disgust & Other   & Total       \\
107      & 40        & 2       & 3       & 7,532       \\
\bottomrule
~\\
\end{tabular}}

\includegraphics[width=0.9\linewidth]{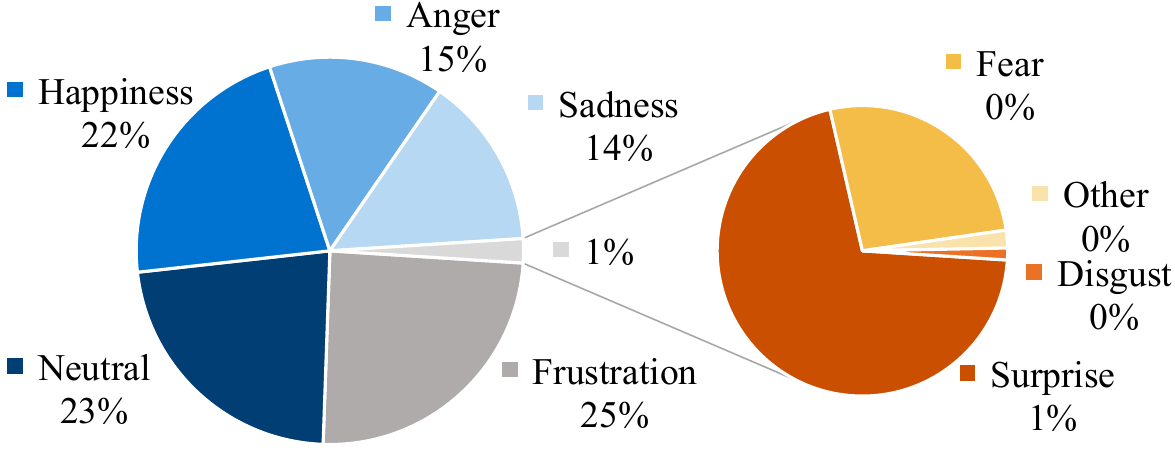}
\label{fig: groud-truth-distr}
\end{table}

As in most previous studies, our 4-way and 5-way classification systems can be evaluated based on classification accuracy with the utterances with majority unique labels. 
However, for the utterances without majority unique labels, that comprise 25\% of IEMOCAP, classification accuracy against a single reference cannot be used\footnote{Multiple reference annotations could be counted as correct for scoring purposes. But it is clearly unsatisfactory when the number of annotators increases while using a small number of emotion classes.}. In fact, the utterances without majority unique labels often include complex and meaningful emotions, and are potentially important for training AER systems. 
Furthermore, similar data will be encountered at test time and the system should also be evaluated on these type of data. In order to understand the problem better, the following section performs a data analysis.

\subsection{Data analysis}
\label{sec: re-exam}
Recall that in IEMOCAP, each utterance was labelled by three annotators and each annotator was allowed to give multiple different labels to each utterance. Table~\ref{tab: example majority unique} shows examples of typical situations for the  hard labels provided by the annotators and Table~\ref{tab: multi-label} summarizes some of the statistics of IEMOCAP. There are 1,272 evaluations that have more than one label, indicating the annotators were uncertain about the emotions when evaluating the utterance. 

When labels from different annotators are considered, all annotators agreed on the same emotion class label  (\textit{e.g.}  `AAA') for only 24\% (2,383 out of 10,309) of the utterances, which we denote as 3/3 agreement or $\Omega_{3/3}$ utterances. The rest of the utterances consist of those with agreement from two annotators, denoted as 2/3 agreement or $\Omega_{2/3}$ utterances (\textit{e.g.}  `AAB'), and those without any agreement, referred to as no agreement or $\Omega_{\leq1/3}$ utterances (\textit{e.g.} `ABC'). Note the case shown in the last row of Table~\ref{tab: example majority unique}. Although the labels have majority `AB', the majority is not unique. Thus, the sentence belongs to $\Omega_{\leq1/3}$.

When majority voting is applied, both 3/3 agreement and 2/3 agreement utterances result in the same majority unique  ground truth label (\textit{e.g.} `A'), despite that an annotator assigned different labels to the 2/3 agreement utterances, which causes a loss of the complexity and uncertainty in emotion annotation. This problem is more severe when there isn't a majority label and they are ignored completely in both training and test. Considering the fact that 51\% of the utterances are 2/3 agreement and 25\% have no majority unique label, the strategy to use majority voting on labels significantly changes the true emotion distribution. 
To resolve these problems, in the next section, we propose representing and modelling the emotion using a distribution rather than a single hard label.

\begin{table}[htbp!]
    \centering
    \caption{Examples of typical situations for IEMOCAP annotations. ``e1'', ``e2'', and ``e3'' are three annotators for an utterance. ``A'' ``B'' ``C'' are three different emotion classes. $\Omega_{3/3}$, $\Omega_{2/3}$, and $\Omega_{\leq1/3}$ refer to the cases that the annotations achieve 3/3, 2/3, and no unique agreement. `AB’ for e2 denotes the second evaluator gave two annotations `A' and `B' to the utterance.}
    \label{tab: example majority unique}
    \begin{tabular}{cccccc}
    \toprule
    e1 & e2 & e3 & Majority & Symbol & Majority Unique  \\
    \midrule
    A  & A  & A  & A        & $\Omega_{3/3}$ & $\sqrt{}$ \\
    A  & A  & B  & A        & $\Omega_{2/3}$ & $\sqrt{}$ \\
    A  & AB & C  & A        & $\Omega_{2/3}$ & $\sqrt{}$ \\
    \midrule
    A  & B  & C  &        & $\Omega_{\leq1/3}$ & $\times$ \\
    A  & AB & BC & AB      & $\Omega_{\leq1/3}$ & $\times$ \\
    \bottomrule
    \end{tabular}
\end{table}

\begin{table}[htbp!]
    \centering
    \caption{Statistics of IEMOCAP. An annotator can perform an ``evaluation'' for an utterance. An evaluation can include more than one labels if the annotator is uncertain about the emotions. }
    \label{tab: multi-label}
    \begin{tabular}{lc}
    \toprule
        Number of total utterances & 10,039\\
        Number of total evaluations & 30,117\\
        Evaluation with more than one label &1,272\\ 
        Utterance with more than three labels &1,186  \\
        Average number of labels per utterance & 3.12 \\
    \midrule
        Number of $\Omega_{3/3}$ utterances & 2,383 \\
        Number of $\Omega_{2/3}$ utterances & 5,149 \\
        Number of $\Omega_{\leq1/3}$ utterances & 2,507 \\
    \bottomrule
    \end{tabular}
\end{table}

\section{Uncertainty Estimation in Emotions}
\label{sec:modelling}
To model the uncertainty caused by subjective emotion perception and evaluation, we propose revising the training target for each utterance from a single one-hot hard label (or categorical or 1-of-$K$ label) 
to a continuous-valued \textit{categorical distribution} over the emotion classes. 
For $\Omega_{\leq1/3}$ utterances, this allows the consideration of scenarios where majority unique labels do not exist (\textit{e.g.} the `ABC' and `AABBC' cases in Table~\ref{tab: example majority unique}). For $\Omega_{2/3}$ utterances, it avoids the problem that not all original hard labels are represented by majority voting (\textit{e.g.} the `AAB' and `AABC' cases in Table~\ref{tab: example majority unique}). To learn such a categorical distribution, 
a frequentist approach is used to obtain a ``soft'' label for training using the MLE of the distribution for each utterance. Next, an alternative Bayesian approach is proposed that estimates a separate Dirichlet prior for each utterance. These two approaches can be combined by interpolating the KL loss with the DPN loss.
Finally, a method to evaluate the performance for uncertainty estimation of the 5-way AER systems is proposed. 

\subsection{Soft labels}
\label{sec:soft label}

Denote a categorical distribution $[\mathrm{p}(\omega_1|\boldsymbol{\mu}),\ldots,\mathrm{p}(\omega_K|\boldsymbol{\mu})]^{\text T}=\boldsymbol{\mu}$
as the emotion distribution of an utterance $\boldsymbol{x}$, where $K$ is the number of emotion classes and $\omega_k$ is the $k$-th class. 
Let $\mathcal{D} = \{\boldsymbol{x}_n, \boldsymbol{\mu}_n^{(1)},\ldots,\boldsymbol{\mu}_n^{(M_n)} \}_{n=1}^N$ be a dataset with $N$ utterances  where $\boldsymbol{x}_n$ is the input features of the $n$-th utterance and $\boldsymbol{\mu}_n^{(1)},\ldots,\boldsymbol{\mu}_n^{(M_n)}$ are its $M_n$ labels provided by all annotators. In this paper, $\boldsymbol{\mu}_n^{(m)}$ is a one-hot vector since hard labels are annotated as in IEMOCAP. 
Such hard labels can be considered as samples drawn from the underlying true emotion distribution $\mathrm{p_{tr}}(\boldsymbol{\mu}|\boldsymbol{x})$. 
For brevity, the subscript $n$ is dropped and the following analysis is based on a single utterance $\boldsymbol{x}$ and its associated $M$ labels $\{\boldsymbol{\mu}^{(1)},\ldots,\boldsymbol{\mu}^{(M)}\}$.

One way to obtain the target emotion distribution $\boldsymbol{\mu}$ for each utterance is to use the MLE:
\begin{align}
\label{eqn: soft MLE}
    \bar{\boldsymbol{\mu}} &= \arg\max_{\boldsymbol{\mu}} \ln\mathrm{p}(\boldsymbol{\mu}^{(1)},\ldots,\boldsymbol{\mu}^{(M)}|\boldsymbol{\mu})=\frac{1}{M}\sum_{m=1}^{M}\boldsymbol{\mu}^{(m)}.
\end{align}
The $k$ th element of $\bar{\boldsymbol{\mu}}$, $\bar{\mu}_{k}=\mathrm{p}(\omega_k|\bar{\boldsymbol{\mu}})$, can be obtained for the MLE as the relative frequency of the $\omega_k$ hard labels: 
\[\bar{\mu}_{k} = {N_k}/{\left(\sum\nolimits_{k^{'}=1}^K N_{k^{'}}\right)},\]
where $N_k$ is the number of occurrences of $\omega_k$ in $\{\boldsymbol{\mu}^{(1)},\ldots,\boldsymbol{\mu}^{(M)}\}$. 
Such an MLE-based distribution is referred to as a soft label,
which consists of the proportion of each emotion class.
For instance, if the three 1-of-3 hard labels of an `AAB' utterance are 
[1,0,0], [1,0,0], and [0,1,0], the soft label is [0.67,0.33,0], whereas the majority unique label obtained by majority voting is [1,0,0].   
This comparison shows that a soft label can preserve some uncertainty information derived from the original hard labels. 
A 5-way classification system introduced in Section~\ref{sec: 5-way classification} can be trained using soft labels instead of majority unique labels by minimising the KL divergence between the soft labels and the predictions, which is abbreviated as a ``soft'' system in this paper. 
Denoting the softmax output of the neural network model as 
$\boldsymbol{y}=[\mathrm{p}(\omega_1|\boldsymbol{x},\boldsymbol{\Lambda}),\ldots,\mathrm{p}(\omega_K|\boldsymbol{x},\boldsymbol{\Lambda})]^{\text T}=f_{\boldsymbol{\Lambda}}(\boldsymbol{x})$,
where $\boldsymbol{\Lambda}$ is the collection of model parameters and   $\boldsymbol{y}$ is the predicted emotion distribution, the soft system training loss is the KL divergence between $\bar{\boldsymbol{\mu}}$ and $\boldsymbol{y}$ given by
\begin{align}
\mathcal{L}_{\text{kl}}&=\text{KL}[\bar{\boldsymbol{\mu}}\|\boldsymbol{y}]=\sum\nolimits_{k=1}^{K}\bar{\mu}_k\ln\frac{\bar{\mu}_k}{\mathrm{p}(\omega_k|\boldsymbol{x},\boldsymbol{\Lambda})}. \label{eqn: soft kl loss}
\end{align}

\subsection{Dirichlet prior network}
\label{sec: dpn}

In Section~\ref{sec:soft label}, the average of the observed labels of an utterance is used as the approximation to the true target emotion distribution.
The MLE converges to the true target distributions when
there is an extremely large number of labels available. 
However, this condition cannot be satisfied in real-world AER since often only a small number of annotators (\textit{i.e.} three for IEMOCAP) can be employed for emotion data labelling due to the task complexity and cost. 
This issue can be alleviated from the Bayesian perspective by introducing a prior distribution for the categorical distribution. 
In this section, we introduce the Dirichlet prior network (DPN) ~\cite{malinin2018predictive,Malinin2019ReverseKT}, a neural network model which models  $\mathrm{p}(\boldsymbol{\mu}|\boldsymbol{x}, {\boldsymbol{\Lambda}})$ by predicting the  parameters of its Dirichlet prior distribution.

\begin{figure}[htb]
\centering
\includegraphics[width=0.8\linewidth]{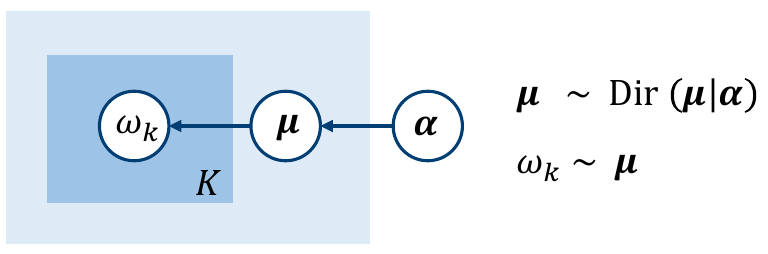}
\caption{Illustration of the DPN process. $\boldsymbol{\mu}$ is a categorical distribution over $K$ emotion classes which is sampled from the Dirichlet prior $\operatorname{Dir}(\boldsymbol{\mu}|{\boldsymbol{{\alpha}}})$.}
\label{fig: DPN}
\end{figure}

The Dirichlet distribution, as the \textit{conjugate prior} of the categorical distribution, is parameterised by its \textit{concentration parameters} $\boldsymbol{{\alpha}}=[{\alpha}_1,\ldots,{\alpha}_K]^{\text T}$.
The Dirichlet distribution $\operatorname{Dir}(\boldsymbol{\mu}| \boldsymbol{{\alpha}})$ is defined as
\begin{align}
    \label{eqn: dirchlet1} \operatorname{Dir}(\boldsymbol{\mu}| \boldsymbol{{\alpha}}) =&\frac{\Gamma\left({\alpha}_{0}\right)}{\prod_{k=1}^{K} \Gamma\left({\alpha}_{k}\right)} \prod_{k=1}^{K} \mu_{k}^{{\alpha}_{k}-1},\\ \nonumber{\alpha}_{0}=&\sum_{k=1}^{K} {\alpha}_{k},{~~}{\alpha}_{k}>0,
\end{align}
where $\Gamma(\cdot)$ is the gamma function defined as
\begin{align}
\Gamma\left({\alpha}_{k}\right)=\int^{\infty}_{0}z^{{\alpha}_{k}-1}e^{-z}\,dz.
\end{align}
Hence, as shown in Fig.~\ref{fig: DPN}, given the concentration parameter $\boldsymbol{{\alpha}}$,
the categorical distribution $\boldsymbol{\mu}$ is a sample drawn from  $\operatorname{Dir}(\boldsymbol{\mu}|\boldsymbol{{\alpha}})$, 
and a 1-of-$K$ hard label relevant to the emotion class $\omega_k$ is a sample drawn from $\boldsymbol{\mu}$.
Here $\boldsymbol{\mu}$ models the distribution over $K$ emotion classes. $\operatorname{Dir}(\boldsymbol{\mu}|{\boldsymbol{{\alpha}}})$ models the distribution of the emotion distribution $\boldsymbol{\mu}$.

For AER, utterance-specific priors derived for each utterance separately are more suitable than a ``global'' prior shared by all utterances, since emotions produced by different speakers in different contexts should not have the same prior. A DPN predicts the concentration parameters ${\boldsymbol{\alpha}}$ as the output of the network ${\boldsymbol{\alpha}}={f}_{{\boldsymbol{\Lambda}}}(\boldsymbol{x})$.
Here the prior distributions are ``utterance-specific'' as they are predicted separately for each utterance.

\subsubsection{Training}
\label{ssec:dpntraining}

Labels provided by different annotators of an utterance can be regarded as categorical distribution samples drawn from the utterance-specific prior.
Given a training utterance $\boldsymbol{x}$ and $M$ categorical emotion distributions $\{\boldsymbol{\mu}^{(1)},\ldots,\boldsymbol{\mu}^{(M)}\}$ provided by the annotators, a DPN is trained to maximise the likelihood $\mathrm{p}(\boldsymbol{\mu} |  \boldsymbol{x}, {\boldsymbol{\Lambda}}) = \operatorname{Dir}(\boldsymbol{\mu} | \boldsymbol{\alpha})$, which is equivalent to minimising the negative log likelihood loss function:
\begin{align}
 \label{eqn: dpn loss}
  \mathcal{L}_{\text{dpn}}
  &=-\frac{1}{M}\sum\nolimits_{m=1}^{M} \ln \mathrm{p}(\boldsymbol{\mu}^{(m)} | \boldsymbol{x},  \boldsymbol{\Lambda})\\
   \nonumber&=-\frac{1}{M}\sum\nolimits_{m=1}^{M}\ln \operatorname{Dir}(\boldsymbol{\mu}^{(m)} | f_{\boldsymbol{\Lambda}}(\boldsymbol{x})),
\end{align}
where $\operatorname{Dir}(\boldsymbol{\mu}^{(m)} | f_{\boldsymbol{\Lambda}}(\boldsymbol{x}))$ is defined in Eqn.~\eqref{eqn: dirchlet1} and $\boldsymbol{\mu}^{(m)}$ is an one-hot hard label. $\mathcal{L}_{\text{dpn}}$ is referred to as the DPN loss in the rest of the paper.

When using the DPN loss, label smoothing~\cite{szegedy2016rethinking} that converts $\boldsymbol{\mu}^{(m)}$ into a ``softer'' label $\hat{\boldsymbol{\mu}}^{(m)}=[\hat{\mu}^{(m)}_1,\ldots,\hat{\mu}^{(m)}_K]$ by
\begin{equation}
	  \hat{\mu}^{(m)}_k=\left\{\begin{array}{ll}
      1- (K-1)\,\varepsilon_1  & \text{if}{~~} \boldsymbol{\mu}^{(m)}\in\omega_k\\
	  \varepsilon_1 & \text{otherwise}
	  \end{array}\right.
\end{equation}
was found necessary to stabilise training, where $\varepsilon_1>0$ is a small constant \cite{malinin2018predictive}. It was also observed that it is important to increase each $\alpha_k$ predicted by the model by another small constant $\varepsilon_2>0$ when calculating $\operatorname{Dir}(\hat{\boldsymbol{\mu}}^{(m)} | f_{\boldsymbol{\Lambda}}(\boldsymbol{x}))$ based on Eqn.~\eqref{eqn: dirchlet1}  \cite{malinin2018predictive}. 

Comparing Eqn. \eqref{eqn: dpn loss} to Eqn. \eqref{eqn: soft kl loss}, each label $\boldsymbol{\mu}^{(m)}$ is taken into account separately when training a DPN, while only the averaged label $\bar{\boldsymbol{\mu}}$ is considered when training a soft label system. For example, two observations `A',`B',`C' and `ABC',`ABC',`ABC' yield the same soft label loss but different DPN losses. The latter case shows that all three annotators are uncertain about the emotion, indicating the emotion of the utterance might have high degree of inherent uncertainty. DPN training preserves the number of occurrences of each emotion class and allows an estimate of the confidence of the uncertainty.

\subsubsection{Inference}
\label{sec: dpn inference}
The predictive distribution of the DPN for an input $\boldsymbol{x}$ is given by marginalising over all possible categorical distributions, which is equivalent to the expected categorical distribution under the conditional Dirichlet prior:
\begin{align}
\nonumber\mathrm{p}(\omega_{k} | \boldsymbol{x}, {\boldsymbol{\Lambda}}) =&\int \mathrm{p}(\omega_{k} |\boldsymbol{\mu}) \  \mathrm{p}(\boldsymbol{\mu} | \boldsymbol{x},{\boldsymbol{\Lambda}}) \mathrm{d} \boldsymbol{\mu}\\
=&\mathbb{E}_{\mathrm{p}(\boldsymbol{\mu} | \boldsymbol{x},{\boldsymbol{\Lambda}})}\left[\mathrm{p}\left(\omega_{k} | \boldsymbol{\mu}\right)\right]
\label{eqn: dpn pred}
\end{align}
which is equal to ${{\alpha}_{k}}/{\sum_{k^{'}=1}^{K} \alpha_{k^{'}}}$ for Dirichlet distribution and is ${\exp({{z}_{k}})}/{\sum_{k^{'}=1}^{K} {\exp}({{z}_{k^{'}}})}$ if an exponential output function $\alpha_k=\exp(z_k)$ is used, where ${z}_{k}$ is the logit of $k$-th output unit of the neural network model. 
This makes the expected posterior probability of an emotion class $\omega_{k}$ be the value of the $k$-th output of the softmax.
In this way, a standard DNN classifier with a softmax output function can be viewed as predicting the expected categorical distribution under a Dirichlet prior~\cite{malinin2018predictive} while the mean is insensitive to arbitrary
scaling of $\alpha_k$. This means that the precision $\alpha_0$, which controls the sharpness of the Dirichlet prior distribution, is degenerate if the classifier is trained with the KL divergence loss instead of the DPN loss.

\subsection{Combining DPN with soft labels}
\label{sec: hybrid}
In this section, we propose to combine the DPN loss $\mathcal{L}_\text{dpn}$ with the KL loss $\mathcal{L}_\text{kl}$ for soft labels by
\begin{equation}
    \label{eqn: total loss}
    \mathcal{L}_{\text{dpn-kl}}=\mathcal{L}_{\text{dpn}}+\lambda\,\mathcal{L}_{\text{kl}},
\end{equation}
where $\lambda$ is a scalar coefficient. Compared to the gradients produced by $\mathcal{L}_{\text{kl}}$, it was observed empirically that those produced by $\mathcal{L}_{\text{dpn}}$ are highly sparse and have a much larger dynamic range. Interpolating $\mathcal{L}_{\text{dpn}}$ with $\mathcal{L}_{\text{kl}}$ reduces the sparsity of the gradients values, and the value of $\lambda$ (\textit{e.g.} 20) is set manually to ensure that the dynamic ranges of the gradients of the $\mathcal{L}_{\text{dpn}}$ and  $\mathcal{L}_{\text{kl}}$ terms are similar.
In practice, it was found in our experiments that using $\mathcal{L}_{\text{dpn-kl}}$ instead of $\mathcal{L}_{\text{dpn}}$ not only stabilised DPN training by removing the need for the two smoothing constants $\varepsilon_1$ and $\varepsilon_2$ defined in Section~\ref{ssec:dpntraining}, but also improved the AER system performance with all evaluation criteria.

Another motivation for using $\mathcal{L}_\text{dpn-kl}$ lies in the connection between $\mathcal{L}_\text{dpn}$ and $\mathcal{L}_\text{kl}$  when an unlimited number of  labels is available for each utterance. Let $\boldsymbol{y}=[\mathrm{p}(\omega_1|\boldsymbol{x},\boldsymbol{\Lambda}),\ldots,\mathrm{p}(\omega_K|\boldsymbol{x},\boldsymbol{\Lambda})]^{\text T}$ be the  distribution of $\boldsymbol{x}$ belonging to each of the $K$ emotion classes estimated by the neural network model, when $M\rightarrow\infty$, the DPN loss can be rewritten as
\begin{equation}
\begin{split}
    \mathcal{L}^{\infty}_\text{dpn}=&-\mathbb{E}_{\mathrm{p}_{\mathrm{tr}}(\boldsymbol{\mu}|\boldsymbol{x})}[\ln \mathrm{p}(\boldsymbol{\mu}|\boldsymbol{x}, \boldsymbol{\Lambda})]\\
=&-\sum\nolimits_{k=1}^{K}\mathbb{E}_{\mathrm{p}_{\mathrm{tr}}(\boldsymbol{\mu}|\boldsymbol{x})}[\mathcal{I}(\boldsymbol{\mu}\in\omega_k) \ln\mathrm{p}(\omega_k|\boldsymbol{x}, \boldsymbol{\Lambda})]\\
=&-\sum\nolimits_{k=1}^{K}\mathrm{p}_{\mathrm{tr}}(\boldsymbol{\mu}\in\omega_k|\boldsymbol{x})\ln\mathrm{p}(\omega_k|\boldsymbol{x}, \boldsymbol{\Lambda})\\
=&\mathrm{KL}\left[\mathrm{p}_{\mathrm{tr}}(\boldsymbol{\mu}|\boldsymbol{x}) \|\boldsymbol{y}\right]+\mathcal{H}\left[\mathrm{p}_{\mathrm{tr}}(\boldsymbol{\mu}|\boldsymbol{x})\right],
\end{split}
\end{equation}
where $\mathcal{I}(\boldsymbol{\mu}\in\omega_k)$ is an indicator function and equals one when $\boldsymbol{\mu}$ is a hard label of $\omega_k$. $\mathcal{H}\left[\mathrm{p}_{\mathrm{tr}}(\boldsymbol{\mu}|\boldsymbol{x})\right]$ is the entropy of $\mathrm{p}_{\mathrm{tr}}(\boldsymbol{\mu}|\boldsymbol{x})$ and is a constant term in the loss. 
$\mathrm{p}_{\mathrm{tr}}(\boldsymbol{\mu}|\boldsymbol{x})$ is the underlying true emotion distribution of $\boldsymbol{x}$,
and $\mathcal{L}^{\infty}_\text{kl}=\mathrm{KL}\left[\mathrm{p}_{\mathrm{tr}}(\boldsymbol{\mu}|\boldsymbol{x}) \|\boldsymbol{y}\right]$ is the KL loss when $M\rightarrow\infty$. Thus both $\mathcal{L}^{\infty}_\text{dpn}$ and $\mathcal{L}^{\infty}_\text{kl}$ reach the same optimum.

With a finite number of labels for each utterance, $\mathcal{L}_\text{dpn}$ and $\mathcal{L}_\text{kl}$ approximate $\mathcal{L}^{\infty}_\text{dpn}$ and $\mathcal{L}^{\infty}_\text{kl}$ separately from different perspectives:
\begin{equation}
\begin{split}
\mathcal{L}^{\infty}_\text{dpn}&\approx-\frac{1}{M}\sum\nolimits_{m=1}^{M}\ln \operatorname{Dir}(\boldsymbol{\mu}^{(m)} | f_{\boldsymbol{\Lambda}}(\boldsymbol{x}))=\mathcal{L}_\text{dpn}\\
\mathcal{L}^{\infty}_\text{kl}&\approx\mathrm{KL}[(\frac{1}{M}\sum\nolimits_{m=1}^M \boldsymbol{\mu}^{(m)}) \| \boldsymbol{y}]=\mathcal{L}_\text{kl}.  \\
\end{split}
\end{equation}
$\mathcal{L}_\text{dpn}$ approximates the expectation with respect to the true distribution by the the empirical average of the likelihood while
 $\mathcal{L}_\text{kl}$ approximates the true distribution by the sample average. When only a finite number of hard labels is available, 
$\mathcal{L}_\text{dpn-kl}$ can achieve a better approximation by leveraging the complementarity of $\mathcal{L}_\text{dpn}$ and $\mathcal{L}_\text{kl}$.
It's worth mentioning that our proposed combined loss could be applicable to other perception and understanding tasks which rely on subjective evaluations and uncertain labels from human annotators.

\subsection{Evaluation of uncertainty estimation}
\label{sec:dpneval}
In the previous sections, a Bayesian approach has been proposed to model the uncertainty in the training labels. However, the most appropriate way to handle the uncertainty in the test data remains an issue. From Table~\ref{tab: multi-label}, at least 25\% of the utterances in IEMOCAP do not have majority unique labels, which makes it impossible to evaluate by classification with a single reference class. This indicates classification accuracy is no longer a suitable evaluation metric when considering more general AER applications, which can encounter test utterances with ambiguous emotional content. Therefore, in this section, we propose using the area under the precision-recall curve (AUPR) with different measures as alternative metrics for AER. 

The AUPR is the average of precision across all recall values computed as the area under the precision-recall (PR) curve. To compute AUPR, a binary task is first defined which, in this paper, is detecting utterances without majority agreed labels (\textit{i.e.} $\Omega_{\leq1/3}$ utterances in IEMOCAP). An uncertainty measure is then defined to measure the uncertainty for each predicted distribution. Two measures are used in this paper:
\begin{itemize}
    \item The probability of the predicted class or max probability (Max.P) that measures the confidence in the prediction  \cite{Lakshminarayanan2017SimpleAS,Lee2018TrainingCC}. Max.P is defined as
\begin{equation}
    \mathcal{P} = \max\nolimits_k \mathrm{p}(\omega_{k}|\boldsymbol{x}, \boldsymbol{\Lambda}).
\end{equation}
\item The entropy of the predictive distribution (Ent.) that has been used in~\cite{Gal2016DropoutAA,Lakshminarayanan2017SimpleAS}. It behaves similarly to Max.P, but represents the confidence encapsulated in the entire output distribution. That is,
\end{itemize}
\begin{equation}
\mathcal{H}\left[\mathrm{p}\left(\boldsymbol{\mu}| \boldsymbol{x}, \boldsymbol{\Lambda}\right)\right]=-\sum\nolimits_{k=1}^{\mathrm{K}} \mathrm{p}\left(\omega_{k} | \boldsymbol{x},\boldsymbol{\Lambda}\right)\ln \mathrm{p}\left(\omega_{k}|\boldsymbol{x}, \boldsymbol{\Lambda}\right).
\end{equation}
A decision threshold is set based on the uncertainty measure which determines whether the test sample belongs to the positive or negative class. For example, utterances with Max.P lower than the threshold (or Ent. higher than the threshold) are predicted as $\Omega_{\leq1/3}$. A PR curve is obtained by calculating the precision and recall for different decision thresholds where the x-axis of a PR curve is the recall, the y-axis is the precision and the decision thresholds are implicit and are not shown as a separate axis. The area under the PR curve is computed as AUPR. Compared to classification accuracy, AUPR can not only be applied to any test utterance but also quantify the model's ability to estimate uncertainty. 

In Section~\ref{sec:modelling exp}, experiments that assess uncertainty estimation ability by detecting utterances without majority agreed labels are reported. This detects whether a test utterance belongs to  $\Omega_{\leq1/3}$ based on the model output distributions with either Max.P or Ent. In the detection experiments, $\Omega_{\leq1/3}$ is chosen as the negative class, while $\Omega_{2/3}$ and $\Omega_{3/3}$ are chosen as the positive classes.
Detecting $\Omega_{\leq1/3}$ was selected as the binary task for AUPR since this simulates such a real application case: if an $\Omega_{\leq1/3}$ utterance is detected, the utterance may include ambiguous emotions that should be evaluated by further models or humans. Otherwise the utterance belongs to $\Omega_{2/3}$ or $\Omega_{3/3}$ where a majority unique label exists and emotion classification can be applied.

\section{Experimental Setup}
\label{sec:expsetup}
Our experimental setup is given in this section, including feature representations and model structures. 

\subsection{Feature representations}
\label{sec:method}

The audio representation used for speech-based AER often includes log-mel filterbank features (FBKs) \cite{Busso2007}. In this paper, 40-dimensional (-d) FBK with a 10~ms frame shift and 25 ms frame length are used, which is denoted FBK$_{25}$. An additional type of long-term FBK feature is also used, which extracted in the same way as FBK$_{25}$ apart from an long 250 ms frame length, and this is denoted FBK$_{250}$.  
FBK features contain information about the short-term spectrum but don't explicitly contain pitch information that can be important in describing emotional speech \cite{pitch-emotion} and is often complementary to FBK features \cite{Dumouchel2009CepstralAL,Pappagari2020}. Following \cite{pitch2014}, log pitch frequency features with probability-of-voicing-weighted mean subtraction over a 1.5 second window are used along with FBK features.

Text features are also included in our models.
Pre-trained 50-d GloVe embeddings are used to encode word-level transcriptions \cite{Glove}, while the pre-trained BERT-based model without fine-tuning is used to encode the transcription of each single utterance into a 768-d vector \cite{devlin-etal-2019-bert}. Following prior work~\cite{tripathi2018multimodal,Majumder2018,Poria2018}, the reference transcriptions from IEMOCAP were used for the text modality.

\subsection{Model structure}
\label{sec:tsbtab}
\begin{figure}[tb] 
\centering    
\includegraphics[width=0.85\linewidth]{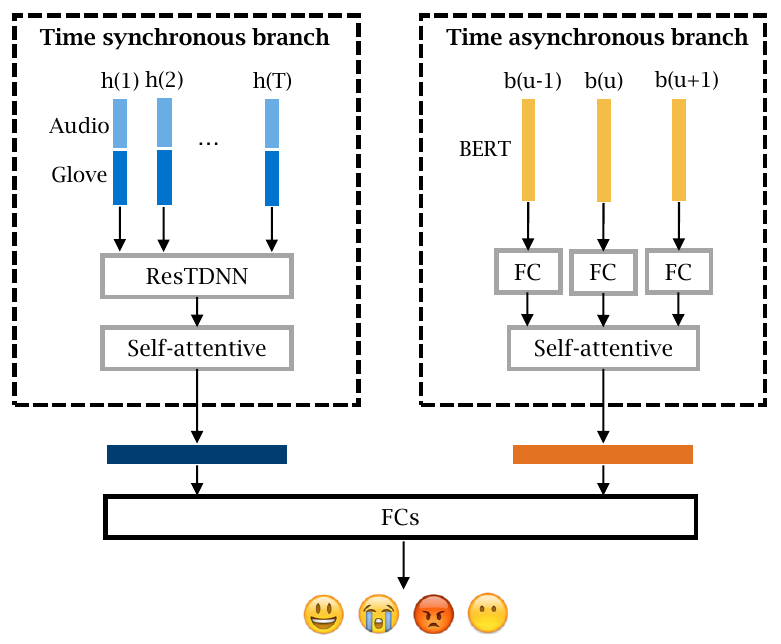}
\caption{Proposed two-branch model structure.}
\vspace{-2mm}
\label{fig:final}
\end{figure}

The proposed model structure is shown in Fig.~\ref{fig:final}, which consists of a time synchronous branch (TSB) that fuses the audio features with the corresponding text information at each time step, and a time asynchronous branch (TAB) that captures the  text information embedded across the transcriptions of a number of consecutive utterances.
In the TSB, the audio features and the corresponding GloVe-based word embeddings are combined at each time step with a simple concatenation operation. The TSB structure is similar to that which is often used for speaker embedding extraction~\cite{Sun2019}. It uses a five-head self-attentive layer \cite{Lin2017ASS} to pool the frame-level vectors across time in the input window, and a time delay neural network with residual connections \cite{Kreyssig2018ImprovedTU} is used as the encoder to derive the frame-level vectors. 

While the TSB includes modelling the temporal correlations between different modalities, the TAB focuses on capturing text information including meaning from the speech transcriptions. The BERT-derived sentence embeddings of the utterance transcriptions are used as the input vectors to the TAB. The embeddings for a number of consecutive utterances were used as the TAB input since the emotion of each utterance is often strongly related to its context in a spoken dialogue~\cite{Mark2012}. 
A shared fully-connected (FC) layer is used to reduce the dimension of each input BERT embedding, and the resulting vectors are then integrated by  
another five-head self-attentive layer.
Finally, output vectors from both branches are fused using an FC layer for emotion classification. The hidden and output activation functions are ReLU and softmax respectively, and a large-margin softmax loss function is used for better regularization~\cite{SphereFace}.

\section{Experiments on IEMOCAP}
\label{sec: classification exp}

Since the test sets are slightly imbalanced between different emotion categories, both the weighted accuracy (WA) and unweighted accuracy (UA) are reported. WA corresponds to the overall accuracy while UA corresponds to the average class-wise accuracy. Models were implemented using HTK~\cite{HTK} in combination with PyTorch. The newbob learning rate scheduler with an initial learning rate of $5\times 10^{-5}$ was used throughout training.

\subsection{4-way classification and cross comparisons}
To compare to previously  published results on IEMOCAP, the system was evaluated with all of the commonly used training and test setups on IEMOCAP: training on Sessions 1-4 and testing on Session 5; 5-fold cross validation (CV) that leaves one of the 5 sessions out of training and used for testing at each fold, and 10-fold CV that leaves one of the 10 speakers out at each fold. These test setups show whether the model is able to learn reliable and speaker independent features with the limited amount of training and test data in IEMOCAP.
The results and modalities used in previous related work are summarised and compared in Table~\ref{tab: liter sum}, which shows that our 4-way classification system achieved state-of-the-art results on IEMOCAP when evaluated with all of the three test settings. More detailed experiments and results can be found in~\cite{wu2021emotion}.

\begin{table}[htbp!]
    \centering
    \caption{Summary of 4-way classification results on IEMOCAP in the literature. ``A'', ``T'', and ``V'' refer to the audio, text, and video modalities respectively. }
    \label{tab: liter sum}
    \begin{tabular}{lcccc}
    \toprule
        Paper & Modality  & Test Setting & WA (\%) & UA (\%)\\
        \midrule
        \cite{tripathi2018multimodal}& A+T+V & Session 5 & 71.04 & -- \\
        \cite{Majumder2018} & A+T+V &  Session 5 & 76.5 & --\\
        \cite{liu20b_interspeech}&A+T & 5-fold CV & 72.39 & 70.08 \\
        \cite{chen20b_interspeech}&A+T & 5-fold CV & 71.06 & 72.05 \\
        \cite{makiuchi2021multimodal} &A+T & 5-fold CV & 73.0 & 73.5 \\
        \cite{Poria2018}& A+T+V & 10-fold CV & 76.1 & --\\
        \midrule
        ours & A+T & Session 5 & \textbf{83.08} & \textbf{83.22} \\
        ours & A+T & 5-fold CV & \textbf{77.57}$\pm 3.89$ & \textbf{78.41}$\pm 3.14$ \\
        ours & A+T & 10-fold CV & \textbf{77.76}$\pm 4.94$ & \textbf{78.30}$\pm 3.97$ \\
        \bottomrule
    \end{tabular}
    
    \vspace{-4mm}
\end{table}

\subsection{5-way classification}
As shown in Table~\ref{tab: 4way5}, the classification accuracy of the 5-way system using 5-fold CV on the previous four emotions (happy, sad, neutral, angry) was 72.47\% WA and 74.29\% UA. A 4.65\% decrease was observed compared with the results of the 4-way system. On the other hand, since the 4-way system cannot correctly classify examples from the ``others'' class, the overall classification accuracy of the 4-way system drops dramatically to 57.02\% WA and 62.72\% UA when tested on the 5-way data. 

\begin{table}[htbp!]
\centering
\caption{Comparison of 4-way and 5-way classification system using 5-fold CV on IEMOCAP.}
\label{tab: 4way5}
\begin{tabular}{ccc}
\toprule
4-way system      & WA(\%)                    & UA(\%)                    \\
\midrule
results of 4 classes & {77.57}$\pm 3.89$ & {78.41}$\pm 3.14$      \\
results of 5 classes & 57.02$\pm 4.23$                     & 62.72$\pm 2.97$ \\
\midrule
\midrule
5-way system      & WA(\%)                    & UA(\%)                    \\
\midrule
results of  4 classes & 72.47$\pm 5.12$ &74.29$\pm 4.90$ \\
results of  5 classes & 77.67$\pm 4.29$ & 77.74$\pm 4.18$\\
\bottomrule
\end{tabular}

\end{table}

\subsection{Uncertainty estimation experiments}
\label{sec:modelling exp}

Four systems were tested: 
\begin{itemize}
    \item ``hard'': A 5-way emotion classification system described in Section~\ref{sec: 5-way classification}; 
    \item ``soft'': A soft label system trained by minimising $\mathcal{L}_{\text{kl}}$ between soft labels and the prediction in Section~\ref{sec:soft label}; 
    \item ``dpn'': A standard DPN system described in Section~\ref{sec: dpn} and trained with  $\mathcal{L}_{\text{dpn}}$ with $\varepsilon_1 = 1
    \times10^{-2}$ and $\varepsilon_2=1
    \times10^{-8}$; 
    \item ``dpn-kl'': A system trained on $\mathcal{L}_{\text{dpn-kl}}$ described in Section~\ref{sec: hybrid} where $\varepsilon_1$ and $\varepsilon_2$ are set to 0, and the weight $\lambda$ to scale the $\mathcal{L}_{\text{kl}}$ term is set to 20.0.  
\end{itemize}
The ``hard'' system was trained on 5-way classification using the training data belonging to $\Omega_{2/3}$ and $\Omega_{3/3}$, while the other systems were all trained using all the utterances in the training set. 
All systems were first evaluated using the 5-way classification accuracy on $\Omega_{2/3}$ and $\Omega_{3/3}$ test utterances, and then evaluated on all test utterances using the average KL divergence, entropy, and AUPR (Max.P) and AUPR (Ent.). 

\begin{table}[htpb]
\centering

\caption{The average of 5-fold CV AER results for uncertainty estimation experiments on IEMOCAP. The WA and UA classification accuracy results were obtained using $\Omega_{\geq 2/3}$ data in the test sets. All the other results were computed on the whole test sets. ``$\uparrow$'' denotes the higher the better, ``$\downarrow$'' denotes the lower the better.} 
\label{tab: dpn-ce}
\begin{tabular}{cccccc}
\toprule
System    & hard    & soft   &  dpn & dpn-kl \\
\midrule
WA(\%)$\uparrow$     & \textbf{77.67}   & 72.29 &72.15 & 75.78  \\
UA(\%)$\uparrow$    & \textbf{77.74}   & 71.34  & 71.19 & 74.76  \\
KL divergence$\downarrow$       & 0.7634  & \textbf{0.5350} & 0.7603 & 0.5580 \\
Entropy    & {0.7065}  & 1.0383 & 1.4467 & 1.0957 \\
AUPR(Max.P)$\uparrow$  & 0.7973 & 0.8230 & 0.8265 & \textbf{0.8468} \\
AUPR(Ent.)$\uparrow$  & 0.7950 & 0.8335 & 0.8316 & \textbf{0.8563} \\
\bottomrule
\end{tabular}

\end{table}

\begin{figure}[htbp!] 
\centering    
\includegraphics[width=0.32\linewidth]{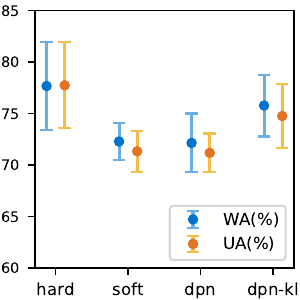}
\includegraphics[width=0.32\linewidth]{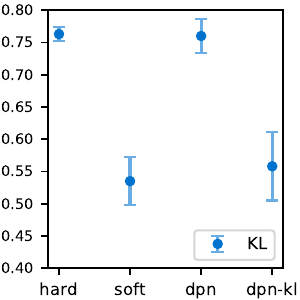}
\includegraphics[width=0.32\linewidth]{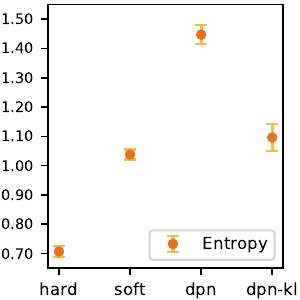}\\

\vspace{2mm}
\includegraphics[width=0.32\linewidth]{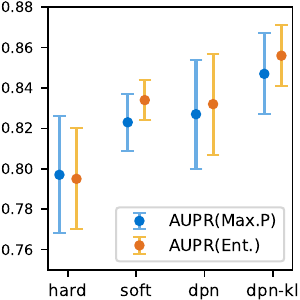}
\includegraphics[width=0.32\linewidth]{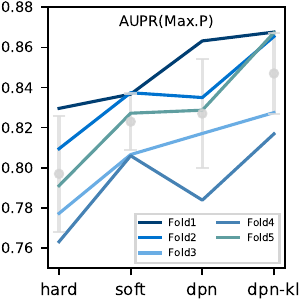}
\includegraphics[width=0.32\linewidth]{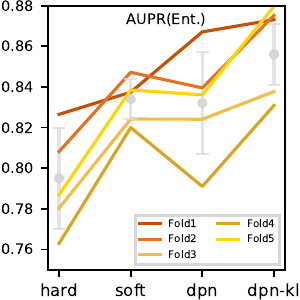}\\
\vspace{-1mm}
\caption{Error bars showing the standard deviation across 5-fold for uncertainty estimation experiments on IEMOCAP. The last two figures show the AUPR value for each fold where ``Fold1'' denotes the 1st fold that was trained on Session 2-5 and tested on Session 1, etc.}
\label{fig: error bar}
\end{figure}

The average of 5-fold CV results on IEMOCAP are shown in Table~\ref{tab: dpn-ce}. Compared to the ``hard'' system that is trained for better emotion classification accuracy, the ``soft'', ``dpn'', and ``dpn-kl'' systems were all trained to better model the uncertainty among the different emotion classes. Therefore it is expected that the hard system has the best UA and WA among all systems. However, as discussed in Section~\ref{sec:dpneval}, classification accuracy is not suitable here as it can not be applied to the $\Omega_{\leq1/3}$ test utterances. It is also expected that the hard system has the lowest entropy (sharper output distributions) as it is trained to learn 0-1 distributions. It is widely known~\cite{nguyen2015deep,hein2019relu} by deep learning researchers that such deep model hard label classification systems are often ``over-confident'', meaning that the model can have poor uncertainty estimation ability as indicated by the AUPR metrics. 
Similarly, it is reasonable that the soft system has the best KL divergence as it is trained to minimise such an  objective. However, the KL divergence is also not the most suitable metric here as it does not distinguish between the `AB' and `AAABBB' cases as discussed in Section \ref{ssec:dpntraining}. Comparing ``dpn'' to ``soft'', ``dpn''  ranks better on AUPR (Max.P) while ``soft'' is better on AUPR (Ent.). 
``dpn-kl'' however, outperforms ``dpn'' on all evaluation metrics and produces the highest AUPR among all systems. It also achieves a balance between the ``hard'' and ``soft'' systems, yielding higher UA and WA than ``soft'' and a smaller KL divergence than ``hard''.  The standard deviation across folds are shown in Fig.~\ref{fig: error bar}. Although the error bars of AUPR value contain overlap among the systems, the ``dpn-kl'' system consistently outperforms the others in all folds.

\subsection{Further experiments for analysis}
For the convenience of visualisation, we took one fold (trained on Session~1-4 and tested on Session~5) as an example and performed further analysis. 
The number of sentences in each group is listed in Table~\ref{tab:soft exp stats}. This section first presents the effect of replacing single categorical hard labels with emotion distributions on the uncertainty of emotion prediction for different data groups by comparing the ``hard'' system to the ``soft'' system. Then the performance of all four systems on detecting the test utterances with high labelling uncertainty is presented.

\begin{table}[htpb]
\centering
\caption{Number of utterances in different data groups of Session 5 of IEMOCAP.}
\label{tab:soft exp stats}
\begin{tabular}{ccccc}
\toprule
    Data group  & $\Omega^{(5)}$ & $\Omega_{3/3}^{(5)}$ & $\Omega_{2/3}^{(5)}$ & $\Omega_{1/3}^{(5)}$\\
\midrule
\# of sentences & 2,170 & 479 & 1,171 & 520\\
\bottomrule
\end{tabular}

\end{table}

\subsubsection{Comparisons between ``hard'' and ``soft'' systems}
\label{ssec:smoothexp}
\begin{figure}[htbp!] 
\centering    
\includegraphics[width=0.9\linewidth]{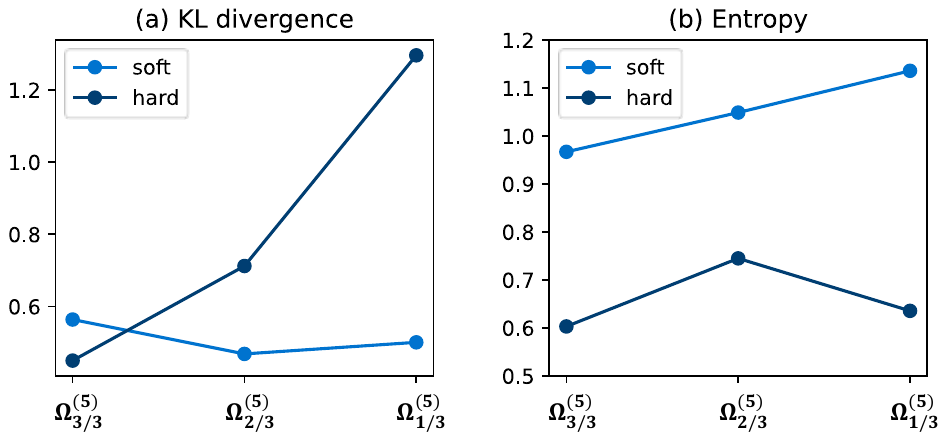}
\caption{Comparison of the ``hard'' system and the ``soft'' system in terms of KL divergence (a) and entropy (b) of three data groups in Session 5 of IEMOCAP.}
\label{fig: soft utt123 compare}
\end{figure}

The performance of the ``hard'' and ``soft'' systems on different data groups are given in Table~\ref{tab:utt-3} and illustrated in Fig.~\ref{fig: soft utt123 compare}. As utterances in $\Omega^{(5)}_{1/3}$ don't have ``ground truth'' hard labels, only the KL divergence and entropy are reported for that group.

\begin{table}[htbp!]
\centering
\caption{Comparison of the ``hard'' system and the ``soft'' system. Tested on different data groups in Session 5 of IEMOCAP.}
\label{tab:utt-3}
\begin{tabular}{ccccc}
\toprule
$\Omega^{(5)}_{3/3}$  & KL divergence & Entropy &  WA(\%) &  UA(\%) \\
\midrule
hard   & 0.4502 & 0.6030 & 87.89      & 87.93  \\
soft   & 0.5640 & 0.9670 & 80.17 & 78.94  \\
\midrule
\midrule
$\Omega^{(5)}_{2/3}$  & KL divergence & Entropy &  WA(\%) &  UA(\%) \\
\midrule
hard   &  0.7122 & 0.7449  & 81.64      & 80.59  \\
soft   & 0.4687 & 1.0489 & 71.90 & 67.50  \\
\midrule
\midrule
$\Omega^{(5)}_{1/3}$  & KL divergence& Entropy&  WA(\%) &  UA(\%) \\
\midrule
hard   &  1.2958 & 0.6355 & -- & -- \\
soft   & 0.5008 & 1.1359 & -- & --\\
\bottomrule
\end{tabular}

\end{table}

As shown in Fig.~\ref{fig: soft utt123 compare}(b),
``soft'' has a higher entropy than ``hard'' in all cases as the soft labels retain some uncertainty in the labels and are trained to produce flatter distributions. The uncertainty in the estimated emotion distribution increases when fewer annotators reach an agreement. The label distribution becomes flatter and the entropy of the distribution predicted by ``soft'' increases. 
Furthermore, as shown in Fig.~\ref{fig: soft utt123 compare}(a), for $\Omega^{(5)}_{3/3}$, which are utterances that all three annotators agree on the same emotion label,  it is easier for ``hard''  to learn the target 0-1 distributions and has  smaller KL divergence. For the $\Omega^{(5)}_{2/3}$ and $\Omega^{(5)}_{1/3}$ utterances, which have more uncertainty in the labels, ``soft'' improves the matching between the distributions and has considerably smaller KL divergence values.

\subsubsection{Evaluating the uncertainty estimation in emotion prediction -- $\Omega_{1/3}$ detection}
\begin{figure}[htbp!]
    \centering
    \includegraphics[width=0.7\linewidth]{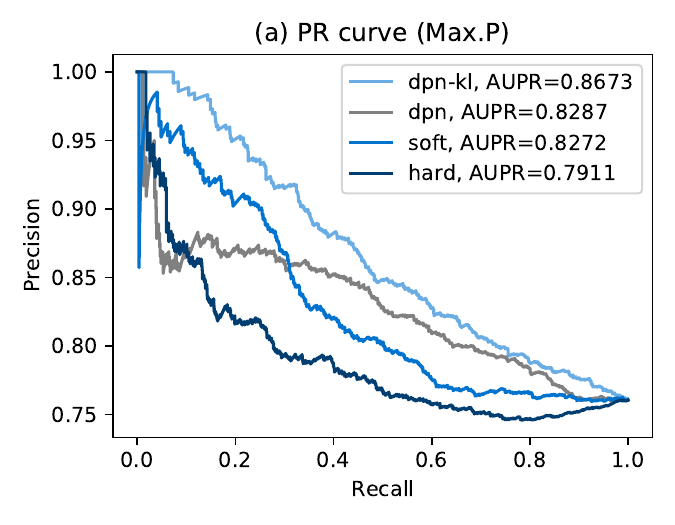}
    \includegraphics[width=0.7\linewidth]{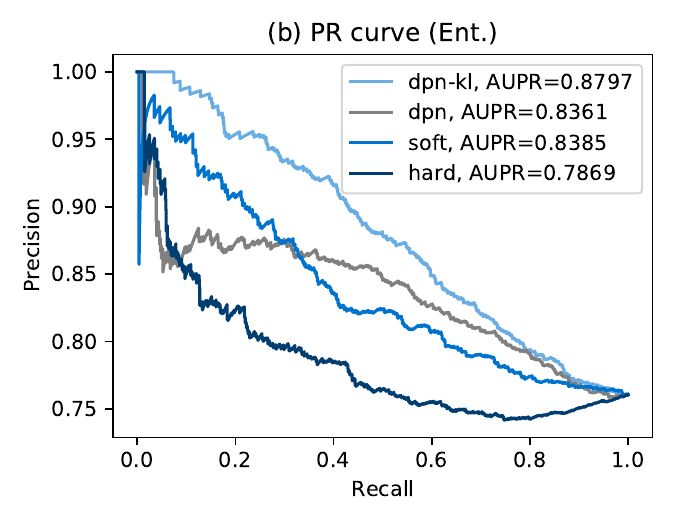}
    \caption{PR curves for the four systems using (a) Max.P and (b) Ent. as the uncertainty measures. The tests were performed on Session 5 of IEMOCAP.}
    \label{fig:AUPR}
\end{figure}

As discussed in Section~\ref{sec:dpneval}, $\Omega_{1/3}$ detection experiments were conducted to assess the models' ability to estimate uncertainty. Both Max.P and Ent. were used as confidence thresholds for AUPR measurement. The precision-recall (PR) curves for all four systems are shown in Fig.~\ref{fig:AUPR}. From the graphs, using Max.P and Ent. as thresholds yield similar trends. The dpn-kl system consistently outperforms all the other systems based on both measures of uncertainty in $\Omega_{1/3}$ detection performance.

\begin{figure}[htbp!]
    \centering
    \includegraphics[width=\linewidth]{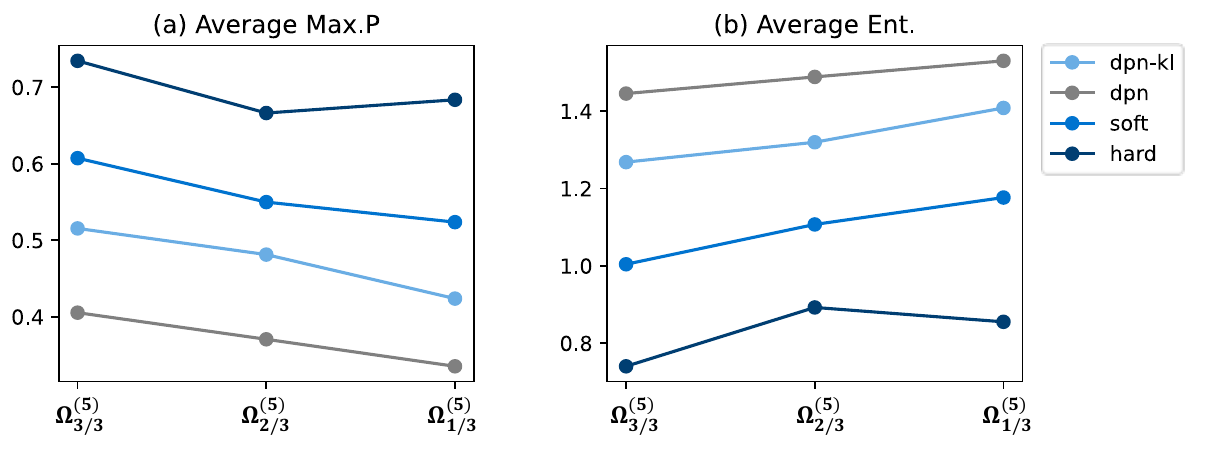}
    \caption{Comparison of average Max.P (a) and Ent. (b) of different data groups in Session 5 of IEMOCAP produced by different systems.}
    \label{fig: avg-maxP-ent ses=5}
\end{figure}

The average value of Max.P and Ent. for different data groups are reported in Table~\ref{tab: avg maxP ent} and illustrated in Fig.~\ref{fig: avg-maxP-ent ses=5}. In general, when the emotion in the utterance gets more complex (with fewer annotators agreeing on the same emotion label), the average Max.P decreases and the average Ent. increases, showing that the systems predicts higher uncertainty of the emotion distribution. The prediction from ``hard'' has the least uncertainty as it is trained with one-hot labels. The high Max.P and low Ent. values of $\Omega_{1/3}$ produced by ``hard'' indicate that this system can give incorrect predictions with high confidence when it encounters test utterances with complex emotions.  
\begin{table}[htbp!]
\centering
\caption{Averaged value of Max.P and Ent. on different test data groups. The tests were performed on Session 5 of IEMOCAP.}
\label{tab: avg maxP ent}
\begin{tabular}{cccc}
\toprule
Averaged Max.P & $\Omega^{(5)}_{3/3}$   & $\Omega^{(5)}_{2/3}$   & $\Omega^{(5)}_{1/3}$  \\
\midrule
hard      & 0.7344 & 0.6662 & 0.6835 \\
soft      & 0.6072 & 0.5498 & 0.5236 \\
dpn       & 0.4053 & 0.3705 & 0.3352 \\
dpn-kl    & 0.5154 & 0.4812 & 0.4237 \\
\midrule
\midrule
Averaged Ent.  & $\Omega^{(5)}_{3/3}$   & $\Omega^{(5)}_{2/3}$   & $\Omega^{(5)}_{1/3}$  \\
\midrule
hard      & 0.7400 & 0.8922 & 0.8550  \\
soft      & 1.0038 & 1.1068 & 1.1764 \\
dpn       & 1.4452 & 1.4882 & 1.5298 \\
dpn-kl    & 1.2679 & 1.3193 & 1.4080 \\
\bottomrule
\end{tabular}

\end{table}

The standard dpn system  exhibits the most uncertainty by producing the lowest Max.P and the highest Ent., indicating that the Dirichlet prior predicted based on only about 3 hard labels introduces a large amount of uncertainty in the estimation. However, such uncertainty plays a key role in the AER problem as it is difficult and expensive to have many reference labels for each utterance and a small number of labels are often insufficient to reflect the true underlying emotion distribution. 
The uncertainty is reduced by smoothing the Dirichlet samples with the MLE by incorporating an additional $\mathcal{L}_{\text{kl}}$  term in the loss function. 
It is worth noting that although ``dpn'' has higher average Ent. than ``dpn-kl'', its AUPR (Ent.) is still worse than ``dpn-kl''.

\subsubsection{Further analysis on emotion data uncertainty}
To understand the influence of the uncertainty in labels, a modified dpn-kl system (referred to as ``dpn-kl2'') was trained using a modified label setting. The labels of each $\Omega_{2/3}$ and $\Omega_{3/3}$ utterance were replaced by the same number of their corresponding majority unique hard label while the hard labels of $\Omega_{1/3}$ utterances were kept the same (referred to as vote-and-replace). An example of this modified label setting is shown in Table~\ref{tab: sum voting eg}, 
and the results are given in Table~\ref{tab: winner-takes-all}.

\begin{table}[b]
\centering
\caption{Example of the vote-and-replace operation of the labels used for dpn-kl2 system. ``A", ``B", ``C'' denotes different emotion categories.}
\label{tab: sum voting eg}
\begin{tabular}{cc}
\toprule
Original label for dpn-kl:          & A A A B C   \\
Majority:  & A \\
Modified label for dpn-kl2:      & A A A A A   \\
\midrule
\midrule
Original label for dpn-kl:          & A B C   \\
Majority:  & ------ \\
Modified label for dpn-kl2:      & A B C  \\
\bottomrule
\end{tabular}

\end{table}
\begin{table}[b]
\centering
\caption{Performance of dpn-kl2 system using labels after modified by the vote-and-replace operation. The tests were performed on Session 5 of IEMOCAP. ``$\uparrow$'' denotes the higher the better, ``$\downarrow$'' denotes the lower the better.}
\label{tab: winner-takes-all}
\begin{tabular}{cccc}
\toprule
System      & hard   & dpn-kl & dpn-kl2 \\
\midrule
WA(\%)$\uparrow$      & 83.45  & 75.21  & 80.91    \\
UA(\%)$\uparrow$     & 82.92  & 72.27  & 79.78    \\
KL divergence$\downarrow$         & 0.7695 & 0.648  & 1.0607    \\
Entropy     & 0.6850  & 1.274  & 0.6621  \\
AUPR(Max.P)$\uparrow$ & 0.7911 & 0.8673 & 0.7434   \\
AUPR(Ent.)$\uparrow$  & 0.7869 & 0.8797 & 0.7385  \\
\bottomrule
\end{tabular}

\end{table}

With vote-and-replace, the UA and WA of ``dpn-kl2''  increase and are closer to those of ``hard''. Compared to ``dpn-kl'', the entropy after voting decreases significantly. The entropy of ``dpn-kl2'' is even smaller than that of ``hard'', possibly because the number of the majority agreed labels 
remain unchanged in ``dpn-kl2'' while only one label is kept by ``hard''.
Taking the first example in Table~\ref{tab: sum voting eg}, the label of ``dpn-kl2'' is `AAAAA' while the label of ``hard'' is `A'.  
Using the same label multiple times to represent the levels of confidence is an advantage of the DPN loss.
Vote-and-replace removes the uncertainty from the $\Omega_{2/3}$ and $\Omega_{3/3}$ utterances. 
This validates our motivation that the majority voting strategy considerably changes the uncertainty properties of the resulting model and should be avoided when constructing AER systems aimed at more general settings.

\section{Experiments on MSP-Podcast}
\label{sec: MSP}
This section presents our experiments on MSP-Podcast~\cite{MSP-podcast}, a larger dataset with natural emotional speech data, to validate the generalisation ability of our proposed method and the reliability of our findings. MSP-Podcast contains natural English speech from podcast recordings. Release 1.8 was used in this paper which contains 73,042 utterances from 1,285 speakers amounting to more than 110 hours of speech. The corpus was annotated using crowd-sourcing. Each utterance was labelled by at least 5 human annotators and has an average of 6.7 annotations per utterance.

\begin{figure}[htbp!]
    \centering
    \includegraphics[width=0.9\linewidth]{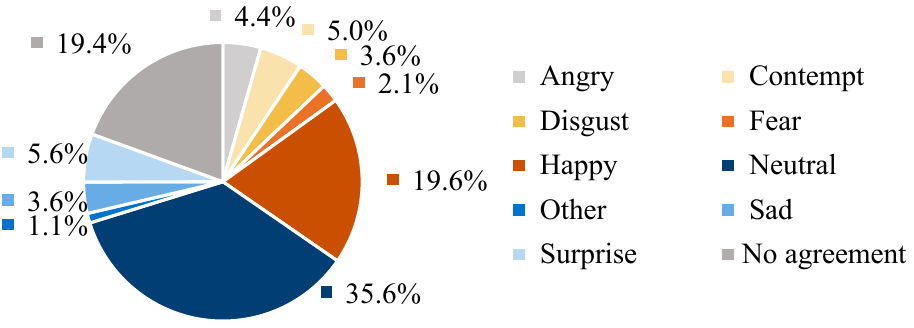}
    \caption{The distribution of emotion classes based on majority agreed labels in MSP-Podcast.}
    \label{fig: podcast-stats}
\end{figure}

The distribution of emotion classes in MSP-Podcast is shown in Fig.~\ref{fig: podcast-stats}. Nearly 20\% of the data doesn't have majority agreed labels, and the distribution is imbalanced among the emotion classes with neutral being the largest emotion class. These are common characteristics of datasets with natural emotions. To counteract the effect of class imbalance, the data was up-sampled by varying the overlap between the input windows. The standard splits for training, validation and testing were used in the experiments.

The TSB-TAB structure was modified to be suitable for MSP-Podcast. Since reference transcriptions were not provided by MSP-Podcast, automatic speech recognition (ASR) results were used instead. We used an open public wav2vec 2.0 large model\footnote{https://huggingface.co/facebook/wav2vec2-large-960h-lv60} to generate the speech transcriptions for MSP-Podcast, which has a word error rate (WER) of 2.2\% on clean test set of Librispeech~\cite{panayotov2015librispeech} (for regular speech) and 39.0\% on IEMOCAP (for emotional speech)\footnote{For comparison, the commercial Google Cloud Speech API has a WER of 44.6\% on IEMOCAP.}. GloVe embeddings were not used here since the end-to-end ASR used here doesn't directly generate word alignments. Furthermore, the BERT-derived sentence embeddings were not used for the context utterances since each emotional utterance is presented separately without providing their surrounding context in MSP-Podcast. The TAB was therefore simplified to an FC layer with ReLU activation. This is a suitable setup for our proposed method that can be used when reference transcriptions and context utterances are not available. 

\begin{table}[htbp!]
\caption{AER results for classification experiments on MSP-Podcast.}
    \centering
    \begin{tabular}{c|ccccc}
    \toprule
    Paper  & Setting  & Modality & WA(\%) & UA (\%) & Weighted F1  \\
    \midrule
    \cite{ando2021speech}  & 4-way & A  & 59.7 & 48.6 & -\\
    \cite{pepino2020fusion}  & 4-way& A+T  &  -& 59.1 & -\\
    \midrule
    \cite{pappagari2021copypaste}  & 5-way & A  & -& -& 0.5868 \\
    \cite{Pappagari2020} & 5-way  & A &- &- & 0.5846 \\
    \midrule
    \cite{chou2022exploiting} & 8-way & A & - & 15.6 & 0.347\\
    \midrule
    ours & 4-way & A+T & 61.85 & 60.63 & 0.6409\\
    ours & 5-way & A+T & 55.43 & 51.97 & 0.6066\\
    ours & 9-way & A+T & 36.93 & 32.84 & 0.3889\\
    \bottomrule
    \end{tabular}
    \label{tab: msp-hard}
\end{table}

Emotion classification results are compared in Table~\ref{tab: msp-hard}\footnote{Note that the results are not directly comparable as different versions of MSP-Podcast dataset were used: papers~\cite{pepino2020fusion,pappagari2021copypaste,Pappagari2020} used release 1.4, papers~\cite{ando2021speech} used release 1.7, and paper~\cite{chou2022exploiting} used release 1.9.} for both our setups and some from the literature. In the 4-way setups, only utterances with majority agreed label belonging to ``happy", ``sad", ``angry", ``neutral'' were used. Emotion class ``disgust'' was included in the 5-way setups. All utterances with majority label were used in the 9-way setup while ``other'' was excluded in the 8-way setup~\cite{chou2022exploiting}. Although our results are not directly comparable to those in the literature as different versions of MSP-Podcast dataset were used, our model produced competitive performance. The degradation of classification results as the number of classes increases indicates the difficulty of fine-grained AER.

The 9-way setup was used in the uncertainty estimation experiments which uses all emotion data. Label grouping was not performed here in order to retain the original labels without modification\footnote{Grouping sentences doesn't affect labelling much in IEMOCAP as in IEMOCAP ``others'' is dominated by ``frustration". }. Emotion was then represented by a 9-dim categorical distribution. The experiments in Table~\ref{tab: dpn-ce} were repeated on MSP-Podcast, and the results shown in Table~\ref{tab: msp-dpn}. 
AUPR was computed by detecting utterances without majority agreed labels. Since each utterance in MSP-Podcast can be labelled by a varying number of annotators, this demonstrates the application of the AUPR metric in a general situation. 
Table~\ref{tab: msp-dpn} shows similar trend to Table~\ref{tab: dpn-ce} with ``hard'' system producing the highest classification accuracy and the ``soft'' system giving the smallest KL divergence. The ``dpn-kl'' system again produces the highest AUPR among all systems, showing its superior ability in emotion uncertainty estimation. Thus we validated the generality of our proposed methods to handle challenging realistic emotion data.

\begin{table}[htbp!]
\centering
\caption{AER results for uncertainty estimation experiments on MSP-Podcast. ``$\uparrow$'' denotes the higher the better, ``$\downarrow$'' denotes the lower the better.}
\label{tab: msp-dpn}
\begin{tabular}{cccccc}
\toprule
System    & hard    & soft   &  dpn & dpn-kl \\
\midrule
WA(\%)$\uparrow$     & \textbf{36.93}   & 33.69 & 33.23 & 35.11  \\
UA(\%)$\uparrow$    & \textbf{32.84}   &  31.67 &  30.06 & {32.68}\\
KL divergence$\downarrow$        & 0.9873 & \textbf{0.8852}  & 1.0834 & 0.9290 \\
Entropy    & 1.6723& 1.9227   &2.1863  &  1.9839 \\
AUPR(Max.P)$\uparrow$  & 0.8101 & 0.8226  & 0.8265& \textbf{0.8437} \\
AUPR(Ent.)$\uparrow$  & 0.8113  & 0.8204  &   0.8278& \textbf{0.8425}\\
\bottomrule
\end{tabular}
\end{table}

\section{Conclusion}
\label{sec:conclusion}

The paper proposes resolving the problem of disagreement in annotated hard labels for emotion classification from the perspective of Bayesian statistics. Instead of using majority voting to achieve a majority hard label that is used to build a classifier, the Dirichlet prior network training loss is applied to the task to better model the distribution of emotions. It preserves label uncertainty by maximising the likelihood of sampling all hard labels with inconsistent emotion classes from an utterance-specific Dirichlet distribution, which is predicted separately for each utterance with a neural network model.
Given the fact that a large proportion of emotion data (\textit{e.g.} in the IEMOCAP dataset) has significant  inter-annotator disagreement, the proposed Bayesian framework also allows the detection of test utterances without majority unique labels based on two uncertainty estimation metrics, 

which is a more general setup than simply ignoring such data as in the traditional emotion classification framework. 
A novel combined loss function that interpolates the DPN loss with Kullback-Leibler loss has also been proposed, which not only has a more stable training performance but also results in improved uncertainty estimates. The findings were further validated using a larger real-life emotion dataset. 
Beyond emotion recognition, label uncertainty is a common issue in many human perception and understanding tasks, since golden references are often not well-defined due to the subjective evaluation of annotators. 
The proposed method could be applicable to other such tasks to handle the uncertainty in labels.

\ifCLASSOPTIONcompsoc
  \section*{Acknowledgments}
\else
  \section*{Acknowledgment}
\fi

Wen Wu is supported by a Cambridge International Scholarship from the Cambridge Trust. This work has been performed using resources provided by the Cambridge Tier-2 system operated by the University of Cambridge Research Computing Service (www.hpc.cam.ac.uk) funded by EPSRC Tier-2 capital grant EP/T022159/1.

The MSP-Podcast data was provided by The University of Texas at Dallas through the Multimodal Signal Processing Lab. This material is based upon work supported by the National Science Foundation under Grants No. IIS-1453781 and CNS-1823166. Any opinions, findings, and conclusions or recommendations expressed in this material are those of the author(s) and do not necessarily reflect the views of the National Science Foundation or The University of Texas at Dallas.

\ifCLASSOPTIONcaptionsoff
  \newpage
\fi



\bibliographystyle{IEEEtran}
\bibliography{ref}

\begin{thebibliography}{10}
\providecommand{\url}[1]{#1}
\csname url@samestyle\endcsname
\providecommand{\newblock}{\relax}
\providecommand{\bibinfo}[2]{#2}
\providecommand{\BIBentrySTDinterwordspacing}{\spaceskip=0pt\relax}
\providecommand{\BIBentryALTinterwordstretchfactor}{4}
\providecommand{\BIBentryALTinterwordspacing}{\spaceskip=\fontdimen2\font plus
\BIBentryALTinterwordstretchfactor\fontdimen3\font minus
  \fontdimen4\font\relax}
\providecommand{\BIBforeignlanguage}[2]{{%
\expandafter\ifx\csname l@#1\endcsname\relax
\typeout{** WARNING: IEEEtran.bst: No hyphenation pattern has been}%
\typeout{** loaded for the language `#1'. Using the pattern for}%
\typeout{** the default language instead.}%
\else
\language=\csname l@#1\endcsname
\fi
#2}}
\providecommand{\BIBdecl}{\relax}
\BIBdecl

\bibitem{Kim2013}
Y.~{Kim}, H.~{Lee}, and E.~M. {Provost}, ``Deep learning for robust feature
  generation in audiovisual emotion recognition,'' in \emph{Proc. ICASSP},
  Vancouver, Canada, 2013, pp. 3687--3691.

\bibitem{Poria2017}
S.~Poria, E.~Cambria, R.~Bajpai, and A.~Hussain, ``A review of affective
  computing: {F}rom unimodal analysis to multimodal fusion,'' \emph{Information
  Fusion}, vol.~37, pp. 98--125, 2017.

\bibitem{Tzirakis2017}
P.~Tzirakis, G.~Trigeorgis, M.~Nicolaou, B.~Schuller, and S.~Zafeiriou,
  ``End-to-end multimodal emotion recognition using deep neural networks,''
  \emph{IEEE Journal of Selected Topics in Signal Processing}, vol.~11, no.~8,
  pp. 1301--1309, 2017.

\bibitem{EMODB}
F.~Burkhardt, A.~Paeschke, M.~Rolfes, W.~F. Sendlmeier, B.~Weiss \emph{et~al.},
  ``A database of {G}erman emotional speech.'' in \emph{Proc. Interspeech},
  Lisbon, Portugal, 2005, pp. 1517--1520.

\bibitem{CHEAVD}
Y.~Li, J.~Tao, L.~Chao, W.~Bao, and Y.~Liu, ``{CHEAVD}: a {C}hinese natural
  emotional audio–visual database,'' \emph{Journal of Ambient Intelligence
  and Humanized Computing}, vol.~8, Sep 2016.

\bibitem{EmoVox}
S.~Albanie, A.~Nagrani, A.~Vedaldi, and A.~Zisserman, ``Emotion recognition in
  speech using cross-modal transfer in the wild,'' in \emph{Proc. ACM
  Multimedia}, Seoul, Korea, 2018, pp. 292--301.

\bibitem{Busso2008IEMOCAPIE}
C.~Busso, M.~Bulut, C.-C. Lee, A.~Kazemzadeh, E.~Provost, S.~Kim, J.~Chang,
  S.~Lee, and S.~Narayanan, ``{IEMOCAP}: {I}nteractive emotional dyadic motion
  capture database,'' \emph{Language Resources and Evaluation}, vol.~42, pp.
  335--359, 2008.

\bibitem{EmoTV1}
S.~Abrilian, L.~Devillers, S.~Buisine, and J.-C. Martin, ``Emo{TV}1: Annotation
  of real-life emotions for the specification of multimodal affective
  interfaces,'' in \emph{Proc. HCI International}, Las Vegas, Nevada, US, 2005,
  pp. 407--408.

\bibitem{MSP-IMPROV}
C.~Busso, S.~Parthasarathy, A.~Burmania, M.~AbdelWahab, N.~Sadoughi, and E.~M.
  Provost, ``{MSP-IMPROV}: An acted corpus of dyadic interactions to study
  emotion perception,'' \emph{IEEE Trans on Affective Computing}, vol.~8,
  no.~1, pp. 67--80, 2017.

\bibitem{MSP-podcast}
R.~Lotfian and C.~Busso, ``Building naturalistic emotionally balanced speech
  corpus by retrieving emotional speech from existing podcast recordings,''
  \emph{IEEE Trans on Affective Computing}, vol.~10, no.~4, pp. 471--483, 2019.

\bibitem{CMU-MOSEI}
A.~Bagher~Zadeh, P.~P. Liang, S.~Poria, E.~Cambria, and L.-P. Morency,
  ``Multimodal language analysis in the wild: {CMU}-{MOSEI} dataset and
  interpretable dynamic fusion graph,'' in \emph{Proc. ACL}, Melbourne,
  Australia, 2018, pp. 2236--2246.

\bibitem{Meld}
S.~Poria, D.~Hazarika, N.~Majumder, G.~Naik, E.~Cambria, and R.~Mihalcea,
  ``{MELD}: A multimodal multi-party dataset for emotion recognition in
  conversations,'' in \emph{Proc. ACL}, Florence, Italy, 2019, pp. 527--536.

\bibitem{devillers2005challenges}
L.~Devillers, L.~Vidrascu, and L.~Lamel, ``Challenges in real-life emotion
  annotation and machine learning based detection,'' \emph{Neural Networks},
  vol.~18, no.~4, pp. 407--422, 2005.

\bibitem{malinin2018predictive}
A.~Malinin and M.~Gales, ``Predictive uncertainty estimation via prior
  networks,'' in \emph{Proc. NeurIPS}, Montréal, Canada, 2018.

\bibitem{wu2021emotion}
W.~Wu, C.~Zhang, and P.~Woodland, ``Emotion recognition by fusing time
  synchronous and time asynchronous representations,'' in \emph{Proc. ICASSP},
  Toronto, Canada, 2021, pp. 6269--6273.

\bibitem{tomkins1962affect}
S.~Tomkins, \emph{Affect imagery consciousness: Volume I: The positive
  affects}.\hskip 1em plus 0.5em minus 0.4em\relax Springer publishing company,
  1962.

\bibitem{ekman1992facial}
P.~Ekman, ``Facial expressions of emotion: New findings, new questions,''
  \emph{Psychological Science}, vol.~3, pp. 34 -- 38, 1992.

\bibitem{ekman2004emotions}
{P. Ekman}, ``Emotions revealed: recognising facial expressions,'' \emph{BMJ},
  vol.~12, pp. 140--142, 2004.

\bibitem{fehr1984concept}
B.~Fehr and J.~A. Russell, ``Concept of emotion viewed from a prototype
  perspective.'' \emph{Journal of experimental psychology: General}, vol. 113,
  no.~3, p. 464, 1984.

\bibitem{cowen2017self}
A.~S. Cowen and D.~Keltner, ``Self-report captures 27 distinct categories of
  emotion bridged by continuous gradients,'' \emph{Proc. National Academy of
  Sciences}, vol. 114, no.~38, pp. E7900--E7909, 2017.

\bibitem{russell1980circumplex}
J.~A. Russell, ``A circumplex model of affect.'' \emph{Journal of personality
  and social psychology}, vol.~39, no.~6, p. 1161, 1980.

\bibitem{grimm2007primitives}
M.~Grimm, K.~Kroschel, E.~Mower, and S.~Narayanan, ``Primitives-based
  evaluation and estimation of emotions in speech,'' \emph{Speech
  communication}, vol.~49, no. 10-11, pp. 787--800, 2007.

\bibitem{schneirla1959evolutionary}
T.~C. Schneirla, ``An evolutionary and developmental theory of biphasic
  processes underlying approach and withdrawal.'' \emph{Nebraska symposium on
  motivation}, pp. 1--42, 1959.

\bibitem{lang1997motivated}
P.~J. Lang, M.~M. Bradley, and B.~N. Cuthbert, ``Motivated attention: Affect,
  activation, and action,'' \emph{Attention and orienting: Sensory and
  motivational processes}, vol.~97, p. 135, 1997.

\bibitem{watson1999two}
D.~Watson, D.~Wiese, J.~Vaidya, and A.~Tellegen, ``The two general activation
  systems of affect: Structural findings, evolutionary considerations, and
  psychobiological evidence.'' \emph{Journal of personality and social
  psychology}, vol.~76, no.~5, p. 820, 1999.

\bibitem{CHEAVD2.0}
Y.~Li, J.~Tao, B.~Schuller, S.~Shan, D.~Jiang, and J.~Jia, ``{MEC} 2017:
  Multimodal emotion recognition challenge,'' in \emph{Proc. ACII Asia},
  Beijing, China, 2018, pp. 1--5.

\bibitem{RECOLA}
F.~Ringeval, A.~Sonderegger, J.~Sauer, and D.~Lalanne, ``Introducing the
  {RECOLA} multimodal corpus of remote collaborative and affective
  interactions,'' in \emph{Proc. FG13}, Shanghai, China, 2013, pp. 1--8.

\bibitem{han2021exploring}
J.~Han, Z.~Zhang, Z.~Ren, and B.~Schuller, ``Exploring perception uncertainty
  for emotion recognition in dyadic conversation and music listening,''
  \emph{Cognitive Computation}, vol.~13, no.~2, pp. 231--240, 2021.

\bibitem{han2017hard}
J.~Han, Z.~Zhang, M.~Schmitt, M.~Pantic, and B.~Schuller, ``From hard to soft:
  Towards more human-like emotion recognition by modelling the perception
  uncertainty,'' in \emph{Proc. ACM Multimedia}, Mountain View, US, 2017, pp.
  890--897.

\bibitem{dang2017investigation}
T.~Dang, V.~Sethu, J.~Epps, and E.~Ambikairajah, ``An investigation of emotion
  prediction uncertainty using gaussian mixture regression.'' in \emph{Proc.
  Interspeech}, Stockholm, Sweden, 2017, pp. 1248--1252.

\bibitem{dang2018dynamic}
T.~Dang, V.~Sethu, and E.~Ambikairajah, ``Dynamic multi-rater {G}aussian
  mixture regression incorporating temporal dependencies of emotion uncertainty
  using {K}alman filters,'' in \emph{Proc. ICASSP}, Calgary, Canada, 2018, pp.
  4929--4933.

\bibitem{atcheson2019using}
M.~Atcheson, V.~Sethu, and J.~Epps, ``Using {G}aussian processes with {LSTM}
  neural networks to predict continuous-time, dimensional emotion in ambiguous
  speech,'' in \emph{Proc. ACII}, Cambridge, UK, 2019, pp. 718--724.

\bibitem{sridhar2021generative}
K.~Sridhar, W.-C. Lin, and C.~Busso, ``Generative approach using soft-labels to
  learn uncertainty in predicting emotional attributes,'' in \emph{Proc. ACII},
  Conference held virtually, 2021, pp. 1--8.

\bibitem{sridhar2020modeling}
K.~Sridhar and C.~Busso, ``Modeling uncertainty in predicting emotional
  attributes from spontaneous speech,'' in \emph{Proc. ICASSP}, Barcelona,
  Spain, 2020, pp. 4929--4933.

\bibitem{tripathi2018multimodal}
S.~Tripathi, S.~Tripathi, and H.~Beigi, ``Multi-modal emotion recognition on
  {IEMOCAP} dataset using deep learning,'' \emph{arXiv preprint 1804.05788},
  2018.

\bibitem{Majumder2018}
N.~Majumder, D.~Hazarika, A.~Gelbukh, E.~Cambria, and S.~Poria, ``Multimodal
  sentiment analysis using hierarchical fusion with context modeling,''
  \emph{Knowledge-Based Systems}, vol. 161, pp. 124--133, 2018.

\bibitem{Poria2018}
S.~Poria, N.~Majumder, D.~Hazarika, E.~Cambria, A.~Gelbukh, and A.~Hussain,
  ``Multimodal sentiment analysis: {A}ddressing key issues and setting up the
  baselines,'' \emph{IEEE Intelligent Systems}, vol.~33, no.~6, pp. 17--25,
  2018.

\bibitem{Nediyanchath2020}
A.~{Nediyanchath}, P.~{Paramasivam}, and P.~{Yenigalla}, ``Multi-head attention
  for speech emotion recognition with auxiliary learning of gender
  recognition,'' in \emph{Proc. ICASSP}, Barcelona, Spain, 2020, pp.
  7179--7183.

\bibitem{Lotfian_2018}
R.~Lotfian and C.~Busso, ``Predicting categorical emotions by jointly learning
  primary and secondary emotions through multitask learning,'' in \emph{Proc.
  Interspeech}, Hyderabad, India, 2018, pp. 951--955.

\bibitem{Ando_2019}
A.~Ando, R.~Masumura, H.~Kamiyama, S.~Kobashikawa, and Y.~Aono, ``Speech
  emotion recognition based on multi-label emotion existence model,'' in
  \emph{Proc. Interspeech}, Graz, Austria, 2019, pp. 2818--2822.

\bibitem{Ando_2018}
A.~Ando, S.~Kobashikawa, H.~Kamiyama, R.~Masumura, Y.~Ijima, and Y.~Aono,
  ``Soft-target training with ambiguous emotional utterances for {DNN}-based
  speech emotion classification,'' in \emph{Proc. ICASSP}, Brighton, UK, 2018,
  pp. 4964--4968.

\bibitem{Fayek_2016}
H.~Fayek, M.~Lech, and L.~Cavedon, ``Modeling subjectiveness in emotion
  recognition with deep neural networks: Ensembles vs soft labels,'' in
  \emph{Proc. IJCNN}, Vancouver, Canada, 2016, pp. 566--570.

\bibitem{Mower09interpretingambiguous}
E.~Mower, A.~Metallinou, C.~chun Lee, A.~Kazemzadeh, C.~Busso, S.~Lee, and
  S.~Narayanan, ``Interpreting ambiguous emotional expressions,'' in
  \emph{Proc. ACII}, Amsterdam, Netherlands, 2009, pp. 1--8.

\bibitem{liu20b_interspeech}
P.~Liu, K.~Li, and H.~Meng, ``Group gated fusion on attention-based
  bidirectional alignment for multimodal emotion recognition,'' in \emph{Proc.
  Interspeech}, Shanghai, China, 2020, pp. 379--383.

\bibitem{chen20b_interspeech}
M.~Chen and X.~Zhao, ``A multi-scale fusion framework for bimodal speech
  emotion recognition,'' in \emph{Proc. Interspeech}, Shanghai, China, 2020,
  pp. 374--378.

\bibitem{makiuchi2021multimodal}
M.~R. Makiuchi, K.~Uto, and K.~Shinoda, ``Multimodal emotion recognition with
  high-level speech and text features,'' in \emph{Proc. ASRU}, Cartagena,
  Colombia, 2021, pp. 350--357.

\bibitem{Malinin2019ReverseKT}
A.~Malinin and M.~J.~F. Gales, ``Reverse {KL}-divergence training of prior
  networks: Improved uncertainty and adversarial robustness,'' in \emph{Proc.
  NeurIPS}, Vancouver, Canada, 2019.

\bibitem{szegedy2016rethinking}
C.~Szegedy, V.~Vanhoucke, S.~Ioffe, J.~Shlens, and Z.~Wojna, ``Rethinking the
  inception architecture for computer vision,'' in \emph{Proc. CVPR}, Las
  Vegas, Nevada, US, 2016, pp. 2818--2826.

\bibitem{Lakshminarayanan2017SimpleAS}
B.~Lakshminarayanan, A.~Pritzel, and C.~Blundell, ``Simple and scalable
  predictive uncertainty estimation using deep ensembles,'' in \emph{Proc.
  NeurIPS}, Long Beach, US, 2017.

\bibitem{Lee2018TrainingCC}
K.~Lee, H.~Lee, K.~Lee, and J.~Shin, ``Training confidence-calibrated
  classifiers for detecting out-of-distribution samples,'' in \emph{Proc.
  ICLR}, Vancouver, Canada, 2018.

\bibitem{Gal2016DropoutAA}
Y.~Gal and Z.~Ghahramani, ``Dropout as a {Bayesian} approximation: Representing
  model uncertainty in deep learning,'' in \emph{Proc. ICML}, New York City,
  US, 2016, pp. 1050--1059.

\bibitem{Busso2007}
C.~Busso, S.~Lee, and S.~Narayanan, ``Using neutral speech models for emotional
  speech analysis,'' in \emph{Proc. Interspeech}, Antwerp, 2007.

\bibitem{pitch-emotion}
E.~Rodero, ``Intonation and emotion: {I}nfluence of pitch levels and contour
  type on creating emotions,'' \emph{Journal of Voice: Official Journal of the
  Voice Foundation}, vol.~25, pp. e25--e34, 2011.

\bibitem{Dumouchel2009CepstralAL}
P.~Dumouchel, N.~Dehak, Y.~Attabi, R.~Dehak, and N.~Boufaden, ``Cepstral and
  long-term features for emotion recognition,'' in \emph{Proc. Interspeech},
  Brighton, UK, 2009, pp. 344--347.

\bibitem{Pappagari2020}
R.~Pappagari, T.~Wang, J.~Villalba, N.~Chen, and N.~Dehak, ``X-vectors meet
  emotions: {A} study on dependencies between emotion and speaker
  recognition,'' in \emph{Proc. ICASSP}, Barcelona, Spain, 2020, pp.
  7169--7173.

\bibitem{pitch2014}
P.~{Ghahremani}, B.~{BabaAli}, D.~{Povey}, K.~{Riedhammer}, J.~{Trmal}, and
  S.~{Khudanpur}, ``A pitch extraction algorithm tuned for automatic speech
  recognition,'' in \emph{Proc. ICASSP}, Florence, Italy, 2014, pp. 2494--2498.

\bibitem{Glove}
J.~Pennington, R.~Socher, and C.~Manning, ``Glo{V}e: {G}lobal vectors for word
  representation,'' in \emph{Proc. EMNLP}, Doha, Qatar, 2014, pp. 1532--1543.

\bibitem{devlin-etal-2019-bert}
J.~Devlin, M.-W. Chang, K.~Lee, and K.~Toutanova, ``{BERT}: {P}re-training of
  deep bidirectional {T}ransformers for language understanding,'' in
  \emph{Proc. NAACL-HLT}, Minneapolis, US, 2019, pp. 4171--4186.

\bibitem{Sun2019}
G.~Sun, C.~Zhang, and P.~Woodland, ``Speaker diarisation using 2{D}
  self-attentive combination of embeddings,'' in \emph{Proc. ICASSP}, Brighton,
  UK, 2019, pp. 5801--5805.

\bibitem{Lin2017ASS}
Z.~Lin, M.~Feng, C.~dos Santos, Y.~Mo, X.~Bing, B.~Zhou, and Y.~Bengio, ``A
  structured self-attentive sentence embedding,'' in \emph{Proc. ICLR}, Toulon,
  France, 2017.

\bibitem{Kreyssig2018ImprovedTU}
F.~Kreyssig, C.~Zhang, and P.~Woodland, ``Improved {TDNNs} using deep kernels
  and frequency dependent {G}rid-{RNNs},'' in \emph{Proc. ICASSP}, Calgary,
  Canada, 2018, pp. 4864--4868.

\bibitem{Mark2012}
F.~Eyben, S.~Buchholz, N.~Braunschweiler, J.~Latorre, V.~Wan, M.~Gales, and
  K.~Knill, ``Unsupervised clustering of emotion and voice styles for
  expressive tts,'' in \emph{Proc. ICASSP}, Kyoto, Japan, 2012, pp. 4009--4012.

\bibitem{SphereFace}
W.~Liu, Y.~Wen, Z.~Yu, M.~Li, B.~Raj, and L.~Song, ``{SphereFace}: {D}eep
  hypersphere embedding for face recognition,'' in \emph{Proc. CVPR}, Hawaii,
  US, 2017, pp. 212--220.

\bibitem{HTK}
S.~Young, G.~E., M.~Gales, T.~Hain, D.~Kershaw, X.~Liu, G.~Moore, J.~Odell,
  D.~Ollason, D.~Povey, A.~Ragni, V.~Valtchev, P.~Woodland, and C.~Zhang,
  \emph{The HTK Book (version 3.5a)}.\hskip 1em plus 0.5em minus 0.4em\relax
  University of Cambridge, 2015.

\bibitem{nguyen2015deep}
A.~Nguyen, J.~Yosinski, and J.~Clune, ``Deep neural networks are easily fooled:
  High confidence predictions for unrecognizable images,'' in \emph{Proc.
  CVPR}, Boston, US, 2015, pp. 427--436.

\bibitem{hein2019relu}
M.~Hein, M.~Andriushchenko, and J.~Bitterwolf, ``Why {R}e{L}u networks yield
  high-confidence predictions far away from the training data and how to
  mitigate the problem,'' in \emph{Proc. CVPR}, Long Beach, US, 2019, pp.
  41--50.

\bibitem{panayotov2015librispeech}
V.~Panayotov, G.~Chen, D.~Povey, and S.~Khudanpur, ``Librispeech: {An} {ASR}
  corpus based on public domain audio books,'' in \emph{Proc. ICASSP}, South
  Brisbane, Australia, 2015, pp. 5206--5210.

\bibitem{ando2021speech}
A.~Ando, R.~Masumura, H.~Sato, T.~Moriya, T.~Ashihara, Y.~Ijima, and T.~Toda,
  ``Speech emotion recognition based on listener adaptive models,'' in
  \emph{Proc. ICASSP}, Toronto, Canada, 2021, pp. 6274--6278.

\bibitem{pepino2020fusion}
L.~Pepino, P.~Riera, L.~Ferrer, and A.~Gravano, ``Fusion approaches for emotion
  recognition from speech using acoustic and text-based features,'' in
  \emph{Proc. ICASSP}, Barcelona, Spain, 2020, pp. 6484--6488.

\bibitem{pappagari2021copypaste}
R.~Pappagari, J.~Villalba, P.~{\.Z}elasko, L.~Moro-Velazquez, and N.~Dehak,
  ``Copy{P}aste: An augmentation method for speech emotion recognition,'' in
  \emph{Proc. ICASSP}, Toronto, Canada, 2021, pp. 6324--6328.

\bibitem{chou2022exploiting}
H.-C. Chou, C.-C. Lee, and C.~Busso, ``Exploiting co-occurrence frequency of
  emotions in perceptual evaluations to train a speech emotion classifier,'' in
  \emph{Proc. Interspeech}, Incheon, Korea, 2022, pp. 161--165.

\end{thebibliography}
%



%



\begin{IEEEbiographynophoto}{Wen Wu}
Wen Wu received the B.E degree from Fudan University and the MPhil degree from University of Cambridge. She is currently a PhD student at University of Cambridge supervised by Prof. Phil Woodland. Her research interests include audio-visual emotion recognition and Bayesian uncertainty estimation.

\end{IEEEbiographynophoto}
\begin{IEEEbiographynophoto}{Dr. Chang Zhang}
Chao Zhang received his BE and MSc degrees in 2009 and 2012 both from the Department of Computer Science and Technology, Tsinghua University, and his PhD degree in 2017 from Cambridge University Engineering Department (CUED). He is currently an Assistant Professor at the Department of Electronic Engineering, Tsinghua University. Before that, he was a Senior Research Scientist at Google, a Research Associate at CUED, and an advisor and speech team co-leader of JD.com. His research interests include spoken language processing, machine learning, and cognitive neuroscience. He has published 70 peer-reviewed speech and language processing papers and received multiple paper awards. He is also a Visiting Fellow at CUED, and an Associate Member of the IEEE Speech and Language Processing Technical Committee.
\end{IEEEbiographynophoto}

\begin{IEEEbiographynophoto}{Dr. Xixin Wu}
Xixin Wu (Member, IEEE) received his B.S. degree from Beihang University, Beijing, China, his M.S. degree from Tsinghua University, Beijing, China, and his Ph.D. degree from The Chinese University of Hong Kong, Hong Kong. He is currently a Research Assistant Professor with the Stanley Ho Big Data Decision Analytics Research Centre, The Chinese University of Hong Kong. Before this, he worked as a Research Associate with the Machine Intelligence Laboratory, Cambridge University Engineering Department. His research interests include speech synthesis and recognition, speaker verification, and neural network uncertainty.
\end{IEEEbiographynophoto}

\begin{IEEEbiographynophoto}{Prof. Philip C. Woodland}
Philip C. Woodland is a Professor of Information Engineering in the Engineering Department, University of Cambridge, Cambridge, U.K., where he is the Head of the Machine Intelligence Laboratory and  a Professorial Fellow of Peterhouse. After working at British Telecom Research Labs for three years, he returned to a Lectureship at Cambridge in 1989 and became a (Full) Professor in 2002. He has published more than 250 papers in the area of speech and language technology with a major focus on speech recognition systems. He has received  a number of best paper awards including for work on speaker adaptation and discriminative training. He is one of the original coauthors of the HTK toolkit and has continued to play a major role in its development. He was a member of the editorial board of Computer Speech and Language (1994–2009) and is currently a member of the editorial board member of Speech Communication. He was a member of the Speech Technical Committee of the IEEE Signal Processing Society from 1999 to 2003. He is a Fellow of the IEEE, the International Speech Communication Association and the Royal Academy of Engineering.
\end{IEEEbiographynophoto}






\end{document}